\documentclass[journal]{IEEEtran}

\usepackage{cite}
\usepackage[pdftex]{graphicx}
\usepackage{color}
\usepackage{amsmath,bm}
\usepackage{subfigure}
\usepackage{url}
\usepackage{booktabs,bigstrut,multirow,stfloats}
\usepackage{algorithm}
\usepackage{algorithmic}

\begin{document}
%
\title{Towards Reducing Severe Defocus Spread Effects for Multi-Focus Image Fusion via an Optimization Based Strategy}
%
%
%

\author{Shuang~Xu,
	Lizhen~Ji,
	Zhe~Wang,
	Pengfei~Li,
	Kai~Sun,
	Chunxia~Zhang, and
	Jiangshe~Zhang

\thanks{This work was supported
	in part by the National Key Research and Development Program of China
	under Grant 2018AAA0102201, and in part by the National Natural Science
	Foundation of China under Grants 11671317, 61976174. \textit{(Corresponding authors: Jiangshe Zhang.)}}

\thanks{Shuang Xu, Lizhen Ji, Kai Sun, Chunxia Zhang, and Jiangshe Zhang are
	with the School of Mathematics and Statistics, Xi’an Jiaotong University, Xi’an
	12480, China (email: shuangxu@stu.xjtu.edu.cn; jlz\_stat@stu.xjtu.edu.cn;  cxzhang@mail.xjtu.edu.cn; skxiaozi@stu.xjtu.edu.cn; jszhang@mail.xjtu.edu.cn).}
\thanks{Zhe Wang is with the Department of Computer Science, University of Virginia,
	Charlottesville, VA 22904 USA (email: zw6sg@virginia.edu).}
\thanks{Pengfei Li is with theUniversalText group, theDepartment of IntelligentTraffic,
	Hikvision, Shanghai 201100, China (email: lipengfei27@hikvision.com).}%
}

%
%

\markboth{ }%
{ }
%



\maketitle

\begin{abstract}
Multi-focus image fusion (MFF) is a popular technique to generate an all-in-focus image, where all objects in the scene are sharp. However, existing methods pay little attention to defocus spread effects of the real-world multi-focus images. Consequently, most of the methods perform badly in the areas near focus map boundaries. According to the idea that each local region in the fused image should be similar to the sharpest one among source images, this paper presents an optimization-based approach to reduce defocus spread effects. Firstly, a new MFF assessment metric is presented by combining the principle of structure similarity and detected focus maps. Then, MFF problem is cast into maximizing this metric. The optimization is solved by gradient ascent. Experiments conducted on the real-world dataset verify superiority of the proposed model.
\end{abstract}

\begin{IEEEkeywords}
multi-focus image fusion, defocus spread effect, structure similarity
\end{IEEEkeywords}

%
\IEEEpeerreviewmaketitle

\section{Introduction}
\IEEEPARstart{D}{ue} to the limitation of imaging devices and their depth-of-field operation, it is hard to acquire all-in-focus images \cite{Review}. In general, only one plane scene stays in focus and others not in focus are blurred. Multi-focus image fusion (MFF) is a useful and promising digital image post-processing technique to cope with this problem. It generates an all-in-focus image by integrating complementary information from source images of the same scene taken at different focus distances.

\begin{figure}[h]
	\centering
	\subfigure[{MMFNet}] {\includegraphics[width=1\linewidth]{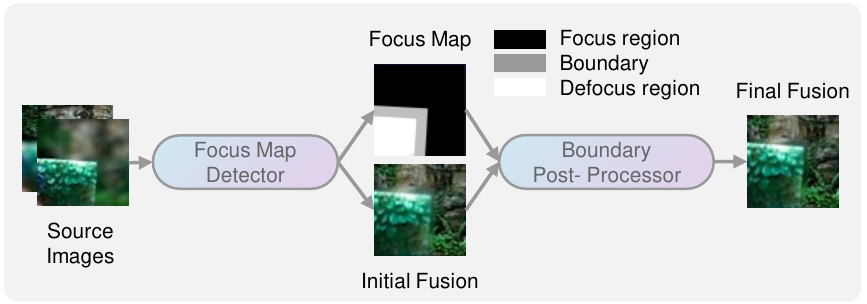}}
	\subfigure[{Our framework}] {\includegraphics[width=1\linewidth]{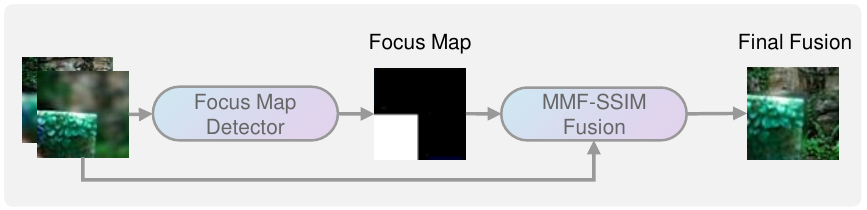}}
	\subfigure[{Fusion image of MMFNet}]{\includegraphics[width=0.49\linewidth]{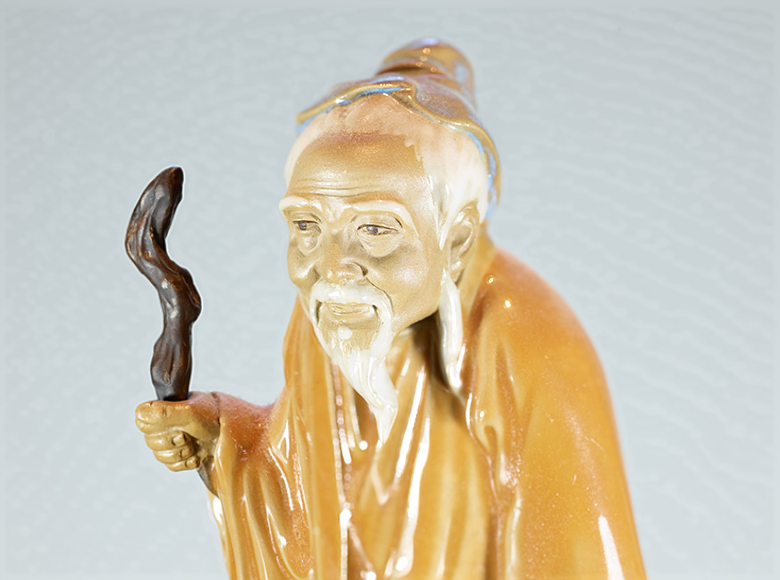}}
	\subfigure[{Fusion image of our framework}]   {\includegraphics[width=0.49\linewidth]{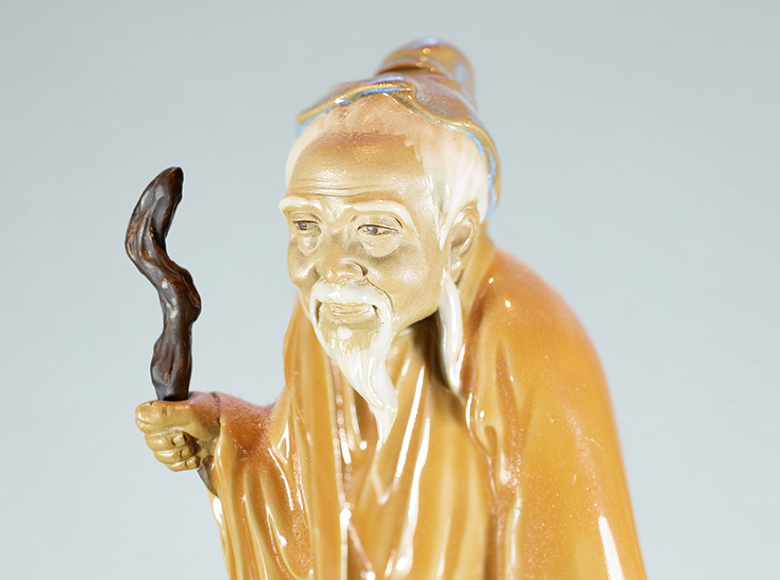}}                   	
	\caption{The differences between MMFNet and our framework. Better view in electronic version.}
	\label{fig:workflow}
\end{figure}

The existing methods can be classified into two groups. The first one is the transform domain based methods. Its basic idea is to utilize a transformer (e.g., discrete Fourier transform\cite{DBLP:journals/access/LiJ20}, discrete wavelet transform \cite{DCT}, non-subsampled contourlet transform \cite{NSCT}, etc.) to convert source images into the feature domain, in which salient features can be easily detected. The fused image is reconstructed from feature domain to spatial domain after merging salient features according to a certain fusion strategy. However, it is reported that transform domain based methods tend to result in the brightness or color distortion because they do not take spatial consistency into account \cite{GFF}. With the development of dictionary learning \cite{SR,DL}, sparse representation based image fusion has emerged as a special transform domain method \cite{SR2}. Sparse representation outperforms classic transforms for its stability and robustness to noise and misregistration \cite{ASR,CSR}. Nonetheless, some details may be lost. Recently, Liu et al. present a general image fusion framework by means of integrating multiscale transform and sparse representation \cite{MST_SR}, which is able to simultaneously overcome their inherent shortcomings. 

\begin{figure*}[!h]
	\centering
	\includegraphics[width=1\linewidth]{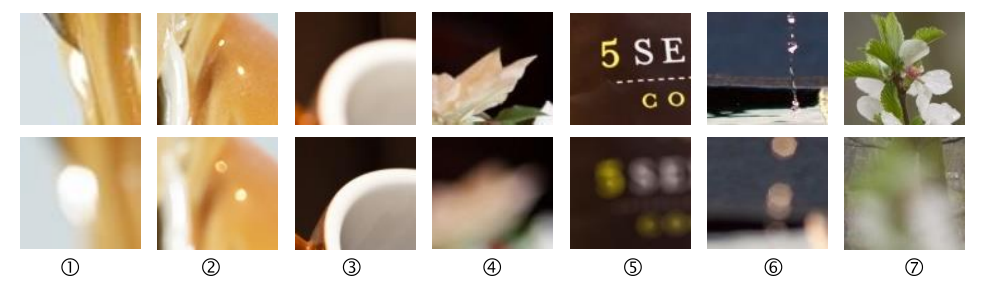}
	\caption{Seven image pairs suffering from defocus spread effects.}
	\label{fig:dse}
\end{figure*}

Spatial domain based methods belong to the second group. They detect the focus map and fuse images in the spatial domain. Generally speaking, how to define a focus measurement and how to accurately detect focus map play significant roles \cite{QIU201935}. To obtain satisfactory results, various sophisticated methods that incorporate certain prior knowledge have been proposed. For example, with the aim at preserving salient edges and local shapes, Li et al. apply a guided filter \cite{GF} to decompose source images into base and detail layers \cite{GFF}. Then, the base and detail layers are fused separately by means of weighted average strategy, where weights are represented by the detected focus map. Finally, the sharp image is reconstructed by combining fused base and detail layers. Li et al. employ the morphological filter to generate initial boundary between focus and defocus regions, and refine it by the matting technique \cite{IM}. However, these methods may lose efficiency when the detected focus map is inaccurate.

Recently, deep neural networks have emerged as effective tools for the MFF task. Liu et al. make the first attempt to apply a convolutional neural network (CNN) to detecting the focus map \cite{Liu_CNN}. Then, they propose a framework for the general image fusion problem \cite{ZHANG202099}. To deal with complicated focus maps, Li et al. design a novel network in the deep regression pair learning (DRPL) fashion \cite{DRPL}. Amin-Naji et al. ensemble the deep features of three neural networks (ECNN) to obtain more accuracy results \cite{ECNN}. Nian and Jung develop a novel CNN to combine the light field data with multi-focus images \cite{DBLP:conf/icip/NianJ19}. The comprehensive comparison of these methods is reported in the recent surveys \cite{LIU202071,DBLP:journals/corr/abs-2005-01116}. Although deep learning based methods are powerful to learn a specific pattern, it is worth pointing out that most of the methods omit defocus spread effects of the real-world multi-focus images \cite{MMFNET,BANET,MFFW,DBLP:journals/corr/abs-2005-01116}. Generally speaking, the out-of-focus objects tend to expand. Hence, when the background object is in focus, the expanded foreground object will overlay the boundary between background and foreground. As a result, many methods are very likely to make mistakes around focus map boundaries and generate unrealistic images, if source images suffer from severe defocus spread effects \cite{MMFNET,BANET,MFFW}. Although several deep learning based methods have been proposed, most of them are devoted to improving the accuracy of focus maps instead of solving the defocus spread effect. 

To the best of our knowledge, there are only two deep networks taking it into account. In references \cite{MMFNET} and \cite{BANET}, they separately proposed two kinds of defocus models to generate synthetic images with defocus spread effects. In addition, they built and trained end-to-end deep neural networks on these synthetic models. The MMFNet proposed in reference \cite{MMFNET} is the state-of-the-art (SOTA) method. As shown in Fig. \ref{fig:workflow} (a), it consists of two sub-networks. The first sub-network serves as a focus map detector who segments the scene into three parts, including a focus region, a defocus region and a focus/defocus boundary. The second sub-network serves as a post-processing network who aims at enhancing the focus/defocus boundary. Nonetheless, the prerequisite of MMFNet performing well is that the focus map detector is accurate enough. Fig. \ref{fig:workflow} (c) exhibits an example, where MMFNet mistakenly detects the focus map and fails to generate clear background. The weak generalization ability and non-robustness limit the application of MMFNet to real-world images. 

To develop a robust fusion strategy, we present a novel optimization-based framework to solve defocus spread effects. The basic idea is to abandon pixel-wise fusion, and to make each local patch of the fused image similar to the corresponding region of the sharpest source image. One of the most suitable image quality metrics is structural similarity (SSIM) \cite{SSIM,MEF_SSIM}, which evaluates the similarity between two images according to the luminance, structure and contrast. However, SSIM cannot be applied to MFF task, because we aim to evaluate the similarity between a fused image and a set of source images rather than a single reference image. To eliminate this obstacle, by combining detected focus maps and the principle of SSIM, we propose the multi-focus image fusion structural similarity (MFF-SSIM) index. Then, MFF-SSIM is taken as an objective function to search for a satisfactory result in the image space. Because MFF-SSIM index is a highly non-linear and non-convex function, it is hard to obtain an analytic solution. As an alternative, the gradient ascent algorithm is employed. Our contributions can be summarized as follows:
\begin{enumerate}
	\item This paper proposes a novel metric called MFF-SSIM index, and the MFF task is turned into maximizing it. An iterative solution of this optimization problem is provided.
	\item A series of experiments are conducted to demonstrate the superiority of MFF-SSIM model. It is revealed that compared with the SOTA techniques, our method is effective to tackle the defocus spread effects which occurs at depth edges in the image in the presence of severe defocus.
\end{enumerate}
The rest of this paper is organized as follows. In section \ref{sec:2}, we present MFF-SSIM index and introduce how to solve our model. Extensive experiments are conducted in section \ref{sec:3}. At last, section \ref{sec:4} concludes the paper.

\section{Model formulation}\label{sec:2}
To begin with, we introduce the notations in this paper. We use the calligraphy letter $\mathcal{X}\in R^{M\times N\times C}$ and the uppercase letter $\bm{X}\in R^{MNC}$ to denote the tensor and vector version of an image, respectively. Here, $M,N$ and $C$ denote the height, width and the number of channels, respectively. The lowercase letter $\bm{x}_i\in R^{CW^2}(i=1,\cdots,P)$ denotes the local patch with window size $W$. The $K$ preregistered multi-focus images and the fused image are denoted by $\{\bm{X}^{[k]}\}_{k=1}^K$ and $\bm{Y}$, respectively. Typically, $\bm{x}_i^{[k]}$ stands for the $i^{\rm th}$ patch of the $k^{\rm th}$ source image. 

\subsection{Defocus spread effects}
As shown in Fig. \ref{fig:dse}, the defocus spread effect is a common phenomenon that the objects not in focus tend to expand. It brings two challenges for multi-focus image fusion algorithms: 
\begin{enumerate}
	\item Some objects not in focus will significantly expand, and they can confuse the focus map detector. As a result, the detection results are inaccurate, and the fused images contain artifacts or inconsistent contents.
	\item Obviously, the blurred foreground will cover a part of clear background, when the background is in focus. On the other hand, the blurred background does not affect the clear foreground, when the foreground is in focus. Thus, there is a region between foreground and background being blurred whenever background or foreground is in focus (e.g., the 6th and the 7th image pairs in Fig. \ref{fig:dse}). 
\end{enumerate}

\subsection{Motivation} 
Recently, segmentation-based methods have emerged as popular tools for MFF task. The basic idea is to detect a focus map $\mathcal{M}$ and then use addition strategy to generate fusion images by the following equation
\begin{equation}\label{eq:1}
\mathcal{Y} = \mathcal{M}\odot\mathcal{X}^{[1]} + (1-\mathcal{M})\odot\mathcal{X}^{[2]},
\end{equation}
where $\odot$ is the element-wise product and $\mathcal{M}$ is binary (its elements equal to 1 if it is in focus and 0 otherwise). This strategy does not consider defocus spread effects into account. To deal with this effect, MMFNet not only generates a focus map $\mathcal{M}$ but also estimate a focus/defocus boundary map $\mathcal{B}$ whose entries indicate whether the pixels are located at the focus/defocus boundary. In other words, MMFNet segments the scene into three parts, i.e., a focus region, a defocus region and the boundary. The fused image is computed by 
\begin{equation}\label{eq:2}
\begin{aligned}
\mathcal{Y} & = (1-\mathcal{B})\odot\left[\mathcal{M}\odot\mathcal{X}^{[1]} 
 + (1-\mathcal{M})\odot\mathcal{X}^{[2]}\right] \\
 & \quad +\mathcal{B}\odot f(\mathcal{X}^{[1]},\mathcal{X}^{[2]}),
\end{aligned}
\end{equation}
where $f(\cdot,\cdot)$ is a boundary post-processing function. However, we found that these two strategies would generate unsatisfactory results if focus maps are inaccuracy or source images suffer from severe defocus spread effects. 

\begin{figure}
	\centering
	\includegraphics[width=1\linewidth]{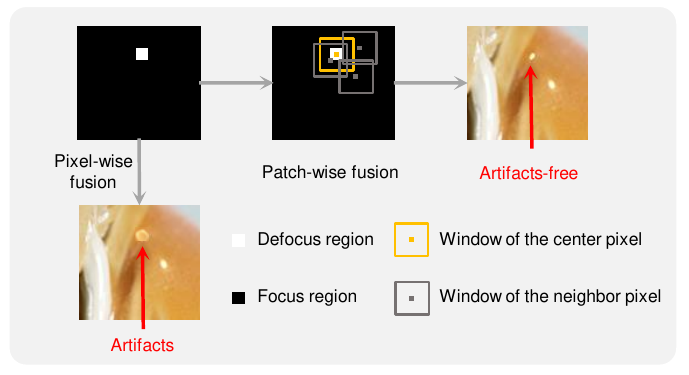}
	\caption{The difference between pixel-wise and patch-wise fusion.}
	\label{fig:motivation}
\end{figure}

In order to reduce the sensitivity to detected focus maps and deal with the defocus spread effect, we present a new framework for the MFF task. It aims at fusing images patch-wise rather than pixel-wise. In intuition, patch-wise fusion may lead to more robust results. Taking the second pair of images in Fig. \ref{fig:dse} as an example, all the pixels in the top patch are in focus. If the detector generates an inaccurate focus map as shown in Fig. \ref{fig:motivation}, the pixel-wise methods would lead to artifacts. Even though these inaccurate maps can be refined by some morphology filters \cite{Liu_CNN}, the experiments reported in subsection \ref{sec:3.2} demonstrate that it still cannot meet the demand (see Fig. \ref{fig:coffee}(b)). The patch-wise method fuses the source images in the patch level. Therefore, when a pixel is mistakenly detected and most of the neighbors in its local patch are correctly detected, this pixel can be corrected by the overlapped windows of its neighbors. According to this idea and inspired by \cite{MEFO,MEF_SSIM}, we propose a novel patch-wise metric to assess the similarity between a fused image and a set of source images in subsection \ref{sec:2-1}. Then, we regard it as the objective function and directly search for the optimal fusion image in the image space. At last, an efficient algorithm is presented to solve this optimization problem in subsection \ref{sec:2-2}.

\subsection{MFF-SSIM index}\label{sec:2-1}
SSIM is a widely used image quality metric. Given two image patches $\bm{x}_i$ and $\bm{y}_i$, SSIM \cite{SSIM} is defined by
\begin{equation}
{\rm SSIM}(\bm{x}_i,\bm{y}_i) = \frac{a_1}{b_1}\frac{a_2}{b_2},
\end{equation}
where 
\begin{equation}\label{eq:mean_var_cov}
\begin{aligned}
a_1&=2\mu_{\bm{x}_i}\mu_{\bm{y}_i}+C_1,& &b_1=\mu_{\bm{x}_i}^2+\mu_{\bm{y}_i}^2+C_1,\\
a_2&=2\sigma_{\bm{x}_i\bm{y}_i}+C_2,&    &b_2=\sigma^2_{\bm{x}_i}+\sigma^2_{\bm{y}_i}+C_2.
\end{aligned}
\end{equation}
Note that $\mu_{\bm{x}_i}$, $\sigma_{\bm{x}_i}^2$ and $\sigma_{\bm{x}_i\bm{y}_i}$ denote the mean, variance and covariance, respectively. $C_1$ and $C_2$ are small constants for numerical stability. Then, the final SSIM score for the two images $\bm{X}$ and $\bm{Y}$ is averaged over all the patches,
$
Q(\bm{X},\bm{Y}) = \frac{1}{P}\sum_{i=1}^{P}{\rm SSIM}(\bm{x}_i,\bm{y}_i),
$
where $P$ denotes the number of patches. SSIM evaluates the similarity between two images, but it unfortunately cannot be directly applied to the MFF task, where we need to assess how much the information is transferred from a set of source images $\{\bm{X}^{[k]}\}_{k=1}^K$ to the fused image $\bm{Y}$. 

To cope with this problem, we introduce the MFF-SSIM index. Generally speaking, a desired fusion image should incorporate the sharper regions of all source images. Hence, MFF-SSIM for the $i^{\rm th}$ patch is defined by 
\begin{equation} \label{eq:4}
S(\{\bm{x}_{i}^{[k]}\}_{k=1}^K,\bm{y}_i)={\rm SSIM}(\bm{x}_{i}^{[j]},\bm{y}_i),
\end{equation}
\begin{center}
	if the $i^{\rm th}$ patch of the $j^{\rm th}$ source image is sharpest,
\end{center}
where $\bm{x}_{i}^{[k]}$ denotes the $i^{\rm th}$ patch of the $k^{\rm th}$ source image. In this fashion, MFF-SSIM is able to compare the fusion image with the sharpest one patch-to-patch. Given the detected focus map, Eq. (\ref{eq:4}) becomes
\begin{equation}
S(\{\bm{x}_{i}^{[k]}\}_{k=1}^K,\bm{y}_i)=\sum_{k=1}^{K}m_{ik}{\rm SSIM}(\bm{x}_{i}^{[k]},\bm{y}_i),
\end{equation}
where $m_{ik}\in \{0,1\}$ indicates whether the $k^{\rm th}$ source image is sharpest with regard to the $i^{\rm th}$ local patch, i.e., the focus map. Then, the final MFF-SSIM is obtained by averaging the local scores, i.e.,
\begin{equation}
Q(\{\bm{X}^{[k]}\}_{k=1}^K,\bm{Y}) = \frac{1}{P}\sum_{i=1}^{P}S(\{\bm{R}_i\bm{X}^{[k]}\}_{k=1}^K,\bm{R}_i\bm{Y}).
\end{equation}
Note that the binary matrix $\bm{R}_i\in \{0,1\}^{CW^2\times MNC}$ is the patch extractor such that $\bm{R}_i\bm{X}=\bm{x}_i$.

\subsection{MFF-SSIM framework}\label{sec:2-2}
Our motivation is to make each local patch of the fused image similar to the corresponding region of the sharpest source image. To this end, the MFF-SSIM index has been proposed to evaluate quality of the fused image in this sense. Therefore, the MFF task can be formulated as the following optimization problem, that is, 
\begin{equation}\label{eq:obj_fun}
\max_{\bm{Y}} Q(\{\bm{X}^{[k]}\}_{k=1}^K,\bm{Y}).
\end{equation}
Owing to the high non-linearity and non-convexity of MFF-SSIM index, obtaining an analytic solution is a challenge. Consequently, we exploit the gradient ascent algorithm to solve this problem. Briefly speaking, given a current estimation of the fused image $\bm{Y}^{(t)}$ and a proper learning rate $\beta>0$, the update 
\begin{equation}\label{eq:ga}
\bm{Y}^{(t+1)} = \bm{Y}^{(t)}+\beta \bm{G}^{(t)}
\end{equation}
will make objective function (MFF-SSIM index) increase. Note that $\bm{G}^{(t)}$ denotes the gradient with regard to (w.r.t.) $\bm{Y}^{(t)}$, $\nabla_{\bm{Y}} Q(\{\bm{X}^{[k]}\}_{k=1}^K,\bm{Y})\arrowvert_{\bm{Y}^{(t)}}$. It is easy to see that 
\begin{equation}
\begin{aligned}
\bm{G}^{(t)}=\frac{1}{P}\sum_{i=1}^P\bm{R}_i^{\rm T}\nabla_{\bm{Y}}S(\{\bm{R}_i\bm{X}^{[k]}\}_{k=1}^K,\bm{R}_i\bm{Y}),
\end{aligned}
\end{equation}
where $\bm{R}_i^T$ denotes the inverse patch extractor to place the gradient patch back into the corresponding entries of the original image. Therefore, the original problem is cast into the computation of gradient for a local patch. We have
\begin{equation}
\begin{aligned}
\nabla_{\bm{y}_i} {\rm SSIM}(\bm{x}_i,\bm{y}_i)&=\frac{a_2\nabla_{\bm{y}_i}a_1+a_1\nabla_{\bm{y}_i}a_2}{b_1b_2} 
\\  &\quad -\frac{a_1a_2(b_2\nabla_{\bm{y}_i}b_1+b_1\nabla_{\bm{y}_i}b_2)}{(b_1b_2)^2},
\end{aligned}
\end{equation}
where 
\begin{equation}
\begin{aligned}
\nabla_{\bm{y}_i}a_1 = 2\mu_{\bm{x}_i}\nabla_{\bm{y}_i}\mu_{\bm{y}_i}&,
\nabla_{\bm{y}_i}a_2 = 2\nabla_{\bm{y}_i}\sigma_{\bm{x}_i\bm{y}_i},\\
\nabla_{\bm{y}_i}b_1 = 2\nabla_{\bm{y}_i}\mu_{\bm{y}_i}&,
\nabla_{\bm{y}_i}b_2 = \nabla_{\bm{y}_i}\sigma_{\bm{y}_i}^2.
\end{aligned}
\end{equation}
The gradient of mean, variance and covariance w.r.t. patch $\bm{y}_i$ are as follows,
\begin{equation}
\begin{aligned}
\nabla_{\bm{y}_i} \mu_{\bm{y}_i} & = \frac{\bm{1}}{CW^2}, \\
\nabla_{\bm{y}_i} \sigma^2_{\bm{y}_i} & =\frac{2(\bm{y}_i-\bm{\mu}_{\bm{y}_i})}{CW^2}, \\
\nabla_{\bm{y}_i} \sigma_{\bm{x}_i\bm{y}_i} &=\frac{(\bm{x}_i-\bm{\mu}_{\bm{x}_i})}{CW^2} .
\end{aligned}
\end{equation}
Here, $\bm{1}$ represents the vector whose all entries are one. According to the above equations, it is easy to write the gradient of a local patch as shown in Eq. (\ref{eq:grdt}). 
\begin{equation}\label{eq:grdt}
\begin{aligned}
& \nabla_{\bm{y}_i}S(\{\bm{x}_i^{[k]}\}_{k=1}^K,\bm{y}_i)\\
=& \frac{2}{CW^2}\sum_{k=1}^{K}m_{ik}\left\lbrace \frac{\mu_{\bm{x}_i^{[k]}}a_2\bm{1}+a_1(\bm{x}_i^{[k]}-\mu_{\bm{x}_i^{[k]}})}
{b_1b_2} \right.\\
&\qquad\qquad\qquad \left. -\frac{a_1a_2[\mu_{\bm{y}_i}b_2\bm{1}+b_1(\bm{y_i}-\mu_{\bm{y}_i})]}
{b_1^2b_2^2} \right\rbrace .
\end{aligned}
\end{equation}
Recall that $m_{ik}\in \{0,1\}$ indicates whether the $i^{\rm th}$ pixel in $k^{\rm th}$ source image is in focus or not. 

\subsection{Implementation details}

Based on the above analysis, we summarize the main steps of MFF-SSIM model as shown in Algorithm \ref{alg:1}. In our experiments, the hyper-parameter configuration is set as follows. When computing the MFF-SSIM index, we set $C_1=0.01^2$ and $C_2=0.03^2$. Furthermore, the overlapped patches are extracted with a stride of 1 to prevent from artifacts around patch boundaries. As for the window size, in section \ref{sec:3.2} it is empirically set as $W=5\times10^{-5}MN$. We also investigate how window size affects our algorithm in section \ref{sec:3.4}. When we optimize MFF-SSIM, the learning rate $\beta$ is set to $10^{-3}$. Our optimization algorithm stops if the number of iterations exceeds 1000. The initial value is set by the average image, that is, $\bm{Y}^{(0)}=\sum_{k=1}^{K}\bm{X}^{[k]}/K$. We rescale images into [0,1]. At last, it should be emphasized that this algorithm needs a pre-detected focus map $\bm{M}$. In the next subsection, we propose two focus map detectors.

The computational complexity of our algorithm is $O(W^2P)$, where $P$ is the number of local patches in an image. Large window size makes our algorithm slow. For a pair of $624\times432$ color images, it takes around 3.23s per iteration with an Intel Core i7-8750H CPU at 2.20GHz.

\begin{algorithm}[t]
	\caption{Gradient ascent for MFF-SSIM model}	\label{alg:1}
	\begin{algorithmic}[1]
		\REQUIRE Initial fused image $\bm{Y}^{(0)}$, learning rate $\beta$, window size $W$, the maximum number of iterations $T$, small constants $C_1,C_2$, the detected focus map $\bm{M}$.
		\ENSURE Final fused image $\bm{Y}^{(t)}$\\ 
		\STATE Compute MFF-SSIM value $Q^{(0)}$ and gradient $\bm{G}^{(0)}$ w.r.t. $\bm{Y}^{(0)}$;
		\FOR {$t=1,2,\cdots,T$}
		\STATE Update fused image $\bm{Y}^{(t)}$ by eqs. (\ref{eq:ga}) and (\ref{eq:grdt});
		\STATE Compute MFF-SSIM value $Q^{(t)}$ and gradient $\bm{G}^{(t)}$ w.r.t. $\bm{Y}^{(t)}$;
		\ENDFOR
	\end{algorithmic}
\end{algorithm}

\begin{figure}
	\centering
	\includegraphics[width=1\linewidth]{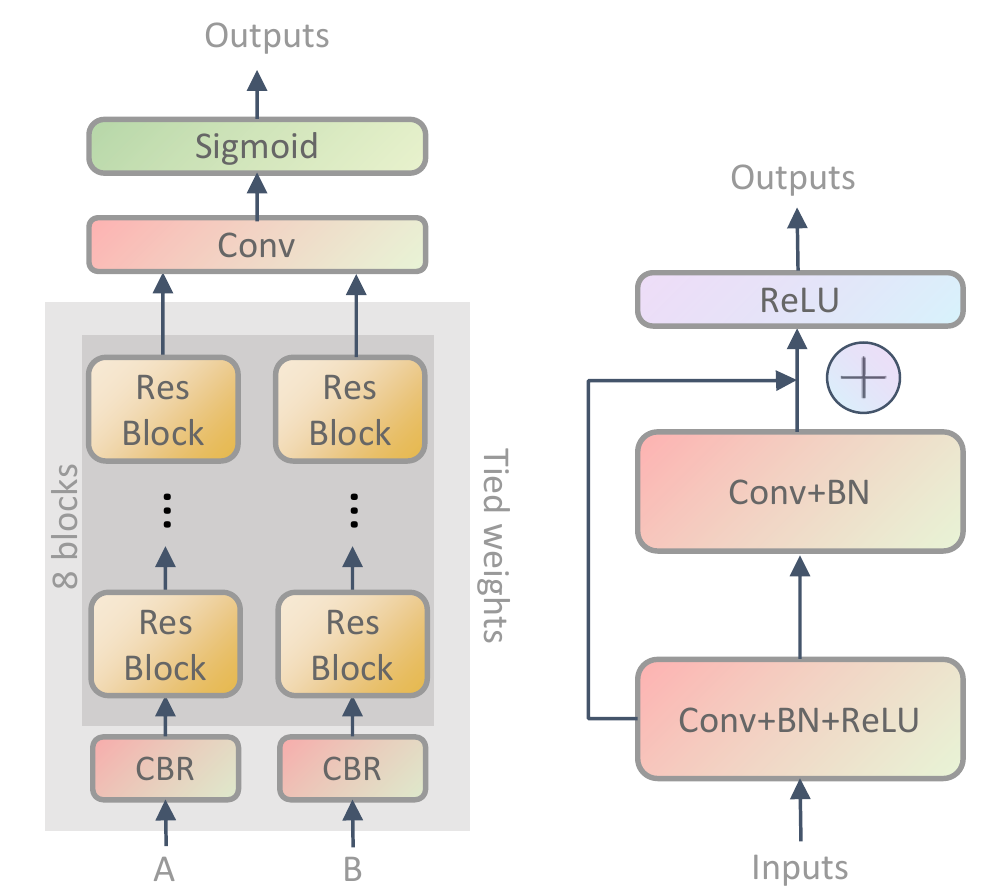}
	\caption{Left: The structure of a residual network. Right: The residual block.}
	\label{fig:structure2}
\end{figure}

\subsection{Focus map detectors}\label{sec:Focus map detectors}
By now, we have proposed the MFF-SSIM framework, but it still lacks a focus map detector. In this paper, we provide two detectors, that is, the Laplacian energy and a residual network. 

According to the fact that a sharp/blurred image has greater/smaller gradients, the focus map can be determined by the gradient intensity. Let $\mathcal{L}$ denote the Laplacian filter, and the gradient intensity of a local patch $\bm{x}_i$ can be quantified by the Laplacian energy, which is defined by
\begin{equation}
	e_{\bm{x}_i} = \sum_{m,n} (\mathcal{L}(\bm{x}_i))_{mn}^2.
\end{equation}
In this way, for the focus map $\bm{M}$, we have $m_{ik}=1$ if $e_{\bm{x}^{[k]}_i}$ is largest among $\{\bm{x}^{[k]}_i\}_{k=1}^{K}$ and 0 otherwise. 

The second detector employs a deep convolutional network to estimate the focus map $\bm{M}$. The network structure is displayed in Fig. \ref{fig:structure2}. At first, the network starts with the CBR (that is, a \textbf{c}onvolutional unit, a \textbf{b}atch normalization (BN) layer and a \textbf{r}ectified linear unit (ReLU)) and 8 residual blocks to separately extract initial feature maps for images A and B. Two feature maps are concatenated along channels and then are put in a convolutional unit and a sigmoid function to generate the focus map. For simplicity, the input images are transformed into gray scale ones. Note that the two sequences of CBR and 8 residual blocks share weights for images A and B. As for the network configuration, there are 128 filters for each convolutional unit except the last one whose number of input and output channels are 256 and 1, respectively. 

We train the network on an image segmentation dataset, PASCAL VOC 2012. The segmentation map is regarded as the ground truth focus map, and the $\alpha$-matte method \cite{MMFNET} is utilized to generate the multi-focus images. Firstly, the clear foreground (${\rm FG^C}$) and background (${\rm BG^C}$) regions are blurred by Gaussian filters and their blurred versions are denoted by ${\rm FG^B}$ and ${\rm BG^B}$. When the foreground is focused, the source image is simulated by the original focus map $\alpha^{\rm C}$. Otherwise, the source image is simulated by the blurred focus map $\alpha^{\rm B}$. In formula, there are 
\begin{equation}
\begin{aligned}
\bm{X}^{[1]} = {\rm FG^C}+(1-\alpha^{\rm C}){\rm BG^B}, \\
\bm{X}^{[2]} = {\rm FG^B}+(1-\alpha^{\rm B}){\rm BG^C}.
\end{aligned}
\end{equation} 
There are 2913 pairs of images in total, so we utilize the image rotation to augment the data. Our network is optimized by Adam over 50 epochs with a batch size of 6 and a learning rate of $10^{-4}$. The loss function is the binary cross-entropy between outputs and ground truth focus maps. 

\begin{table*}[h]
	\centering
	\caption{The results on MFFW dataset. The best and the second best values are highlighted by bold typeface and underline, respectively. } 
	\begin{tabular}{lccccccccc}
		\toprule
		\multirow{2}[4]{*}{Methods} & Objective Metric &       & \multicolumn{3}{c}{Reference Based Metrics} &       & \multicolumn{3}{c}{No Reference Based Metrics} \\
		\cmidrule{2-2}\cmidrule{4-6}\cmidrule{8-10}      & MOS   &       & PSNR  & SSIM  & FSIM  &       & NMI   & Xydeas & Chen-Blum \\
		\midrule
		BF    & 7.3846  &       & 34.0595  & 0.9833  & 0.9809  &       & 1.1104 & 0.5941 & \textbf{0.7456} \\
		BRW   & 7.3846  &       & 35.4858  & \underline{0.9865 } & 0.9839  &       & 1.0415 & \underline{0.6165} & 0.7268 \\
		CBF   & 6.6923  &       & 32.7452  & 0.9741  & 0.9694  &       & 0.8648 & 0.5244 & 0.6554 \\
		GFF   & 8.4615  &       & 35.1882  & 0.9848  & 0.9851  &       & 0.9371 & 0.6022 & 0.7186 \\
		CNN   & 8.5385  &       & 35.1722  & 0.9829  & 0.9831  &       & 1.0638 & \textbf{0.6615} & 0.7362 \\
		ECNN  & 7.9015  &       & 35.0610  & 0.9845  & 0.9829  &       & \textbf{1.1353} & 0.6135 & 0.7321 \\
		DPRL  & 7.3723  &       & 32.8121  & 0.9796  & 0.9751  &       & \underline{1.1214} & 0.6159 & 0.7124 \\
		MMF-Net & 7.3077  &       & 31.8415  & 0.9764  & 0.9709  &       & 0.9174 & 0.4243 & 0.6644 \\
		\midrule
		MS-ResNet & \underline{8.6154 } &       & \textbf{36.4105 } & \textbf{0.9882 } & \textbf{0.9888 } &       & 0.9848 & 0.5858 & 0.7233 \\
		MS-Lap & \textbf{8.9231 } &       & \underline{35.6802 } & 0.9860  & \underline{0.9880 } &       & 1.0151 & 0.6082 & \underline{0.7335} \\
		\bottomrule
	\end{tabular}%
	\label{tab:exp1}%
\end{table*}%

\begin{figure}
	\centering
	\subfigure[{Source 1}] {\includegraphics[width=0.325\linewidth]{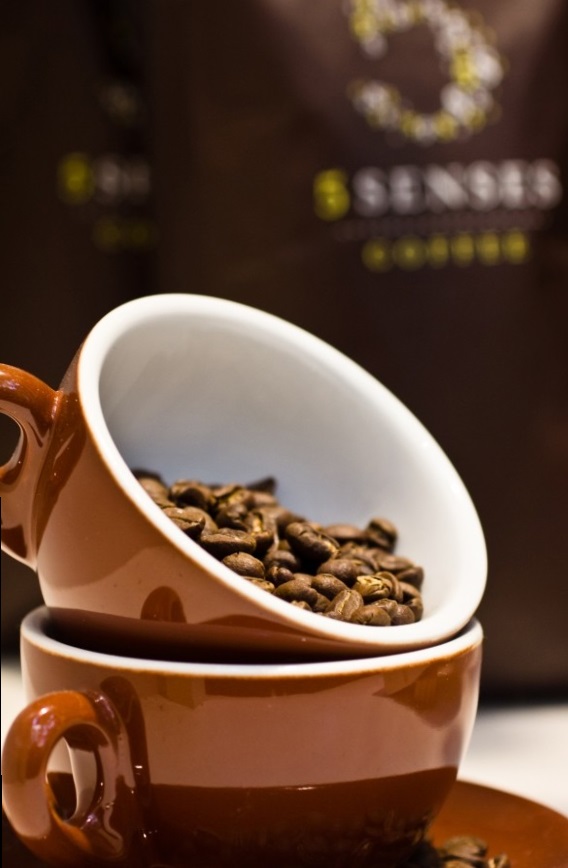}}
	\subfigure[{Source 2}] {\includegraphics[width=0.325\linewidth]{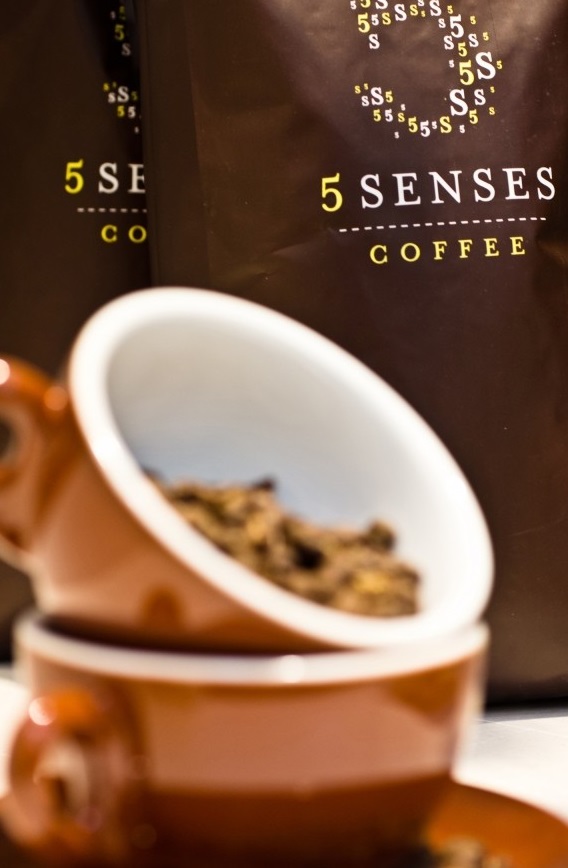}}
	\subfigure[{Reference}]{\includegraphics[width=0.325\linewidth]{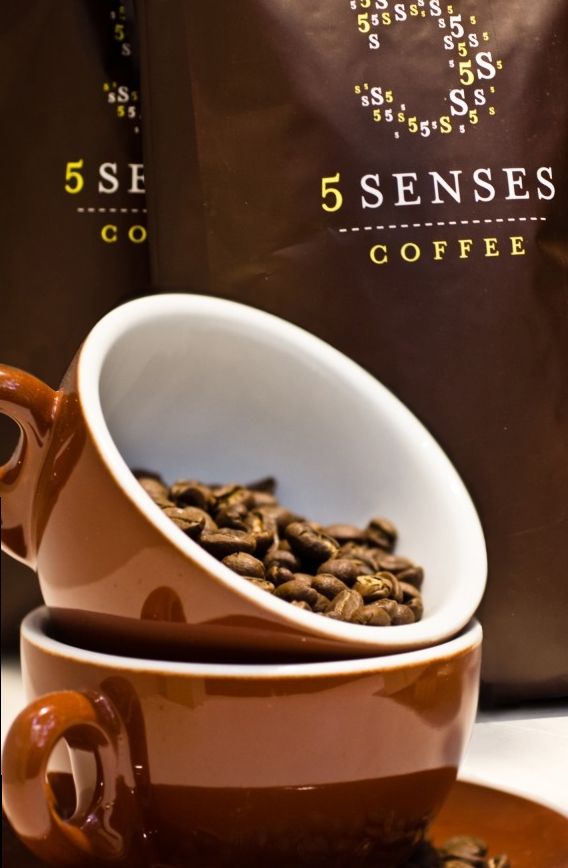}}
	\caption{(a-b) No. 6 image set of the MFFW dataset. (c) The manually fused image. }
	\label{fig:coffee_ori}
\end{figure}
\begin{figure*}
	\centering
	\subfigure[{BF (0.9875)}]                 {\includegraphics[width=0.16\linewidth]{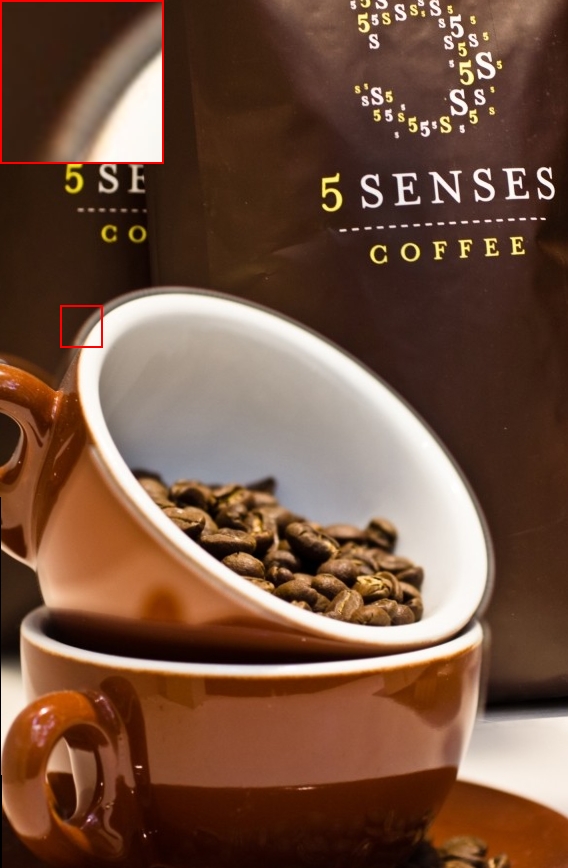}}
	\subfigure[{CNN (0.9240)}]                {\includegraphics[width=0.16\linewidth]{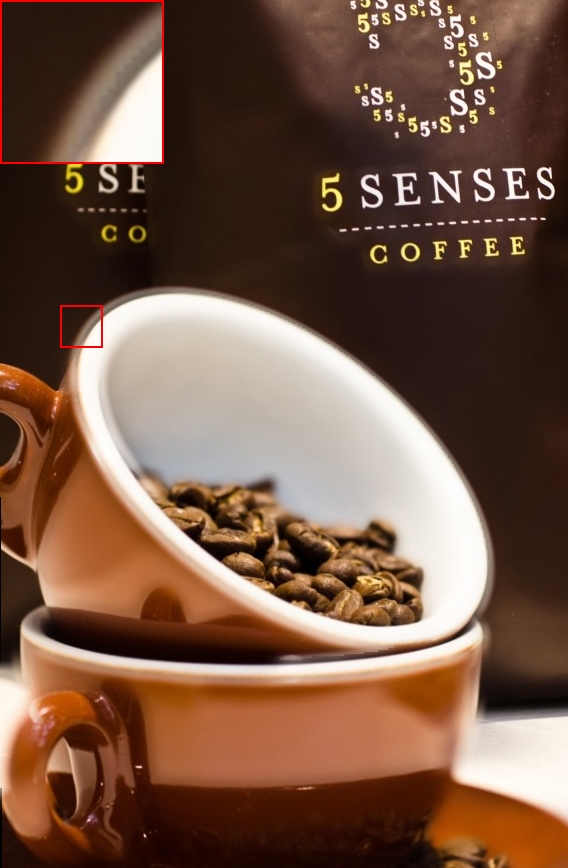}}
	\subfigure[{DPRL (0.9404)}]               {\includegraphics[width=0.16\linewidth]{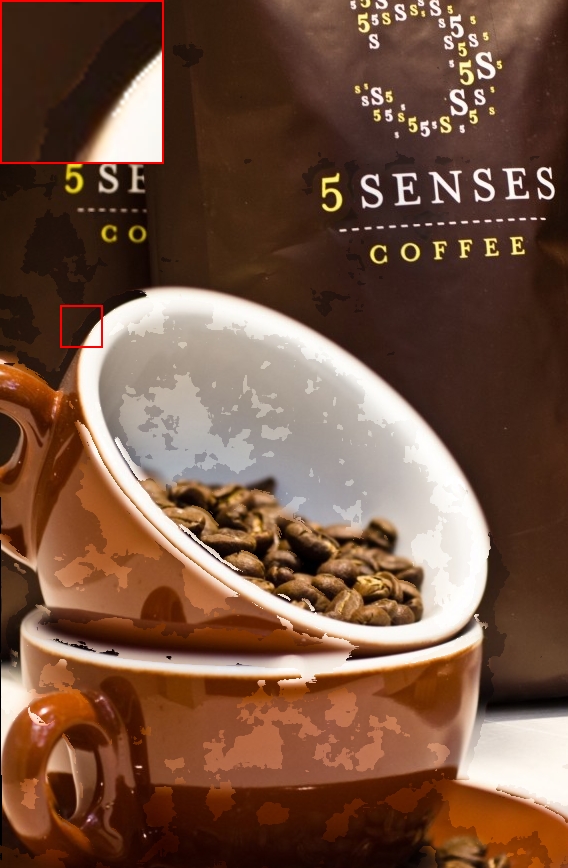}}
	\subfigure[{MMFNet(0.9365)}]              {\includegraphics[width=0.16\linewidth]{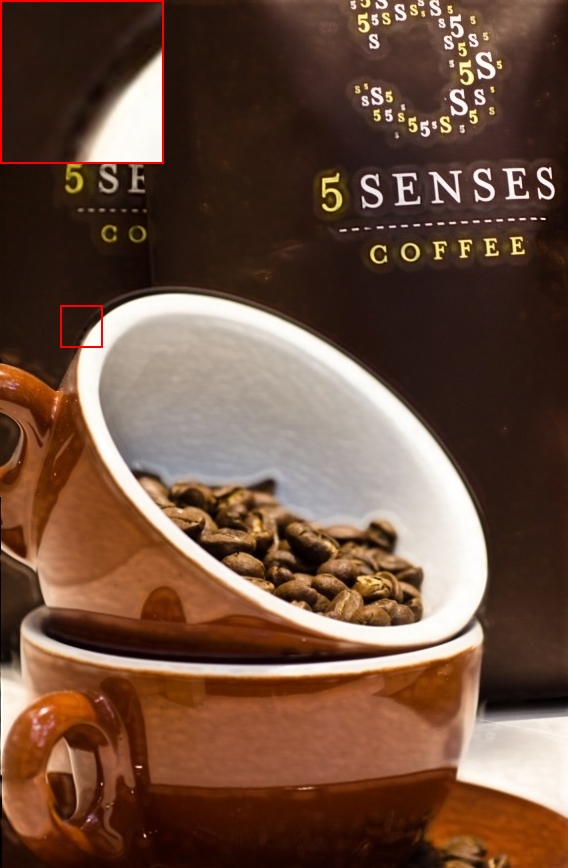}}
	\subfigure[{MS-Lap (\textbf{0.9935})}]   {\includegraphics[width=0.16\linewidth]{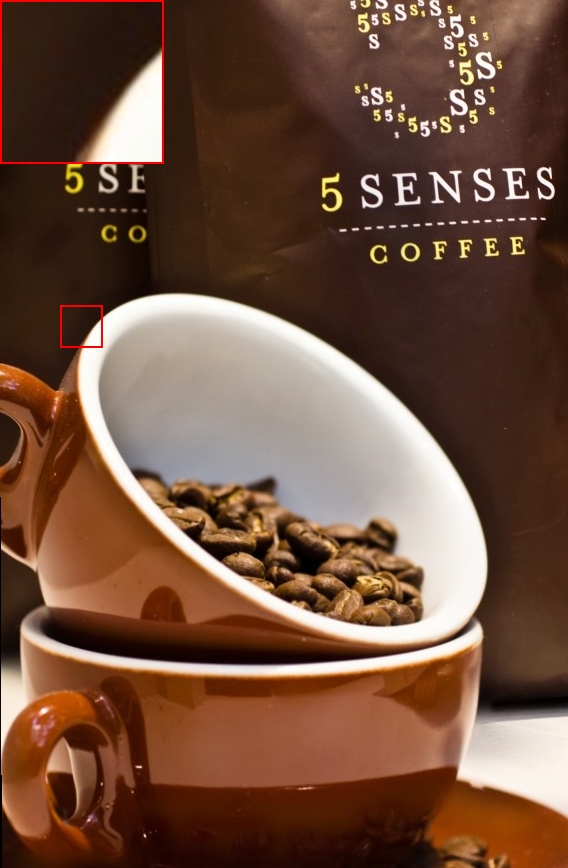}}
	\subfigure[{MS-ResNet (0.9863)}]         {\includegraphics[width=0.16\linewidth]{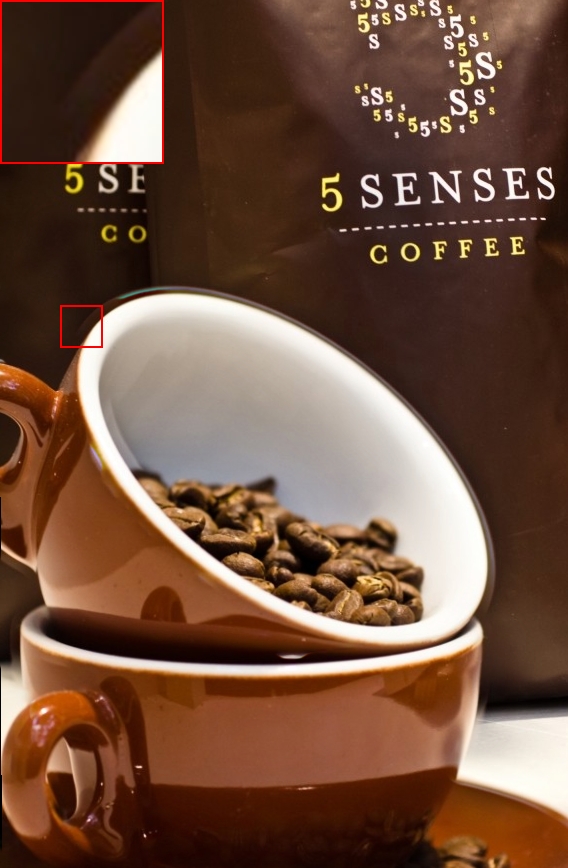}}
	
	\subfigure[{BF Map}]                        {\includegraphics[width=0.16\linewidth]{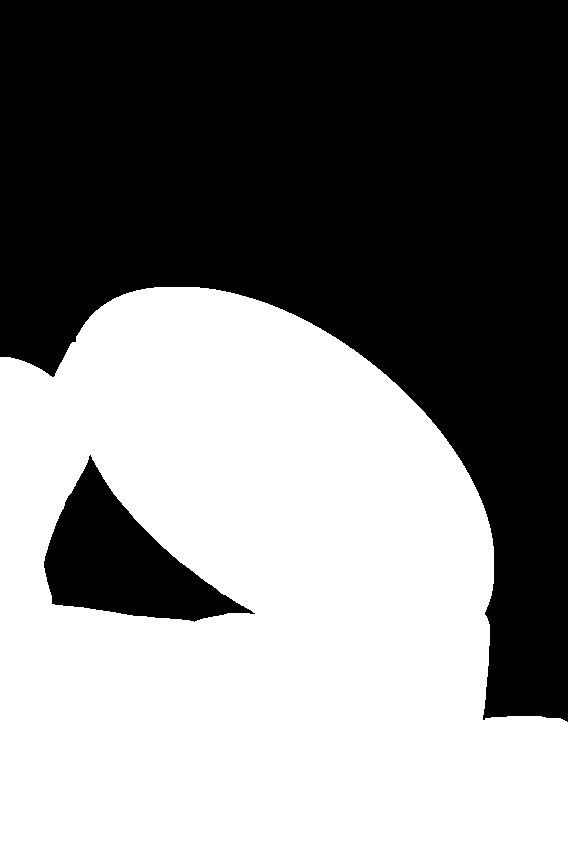}}
	\subfigure[{CNN Map}]                        {\includegraphics[width=0.16\linewidth]{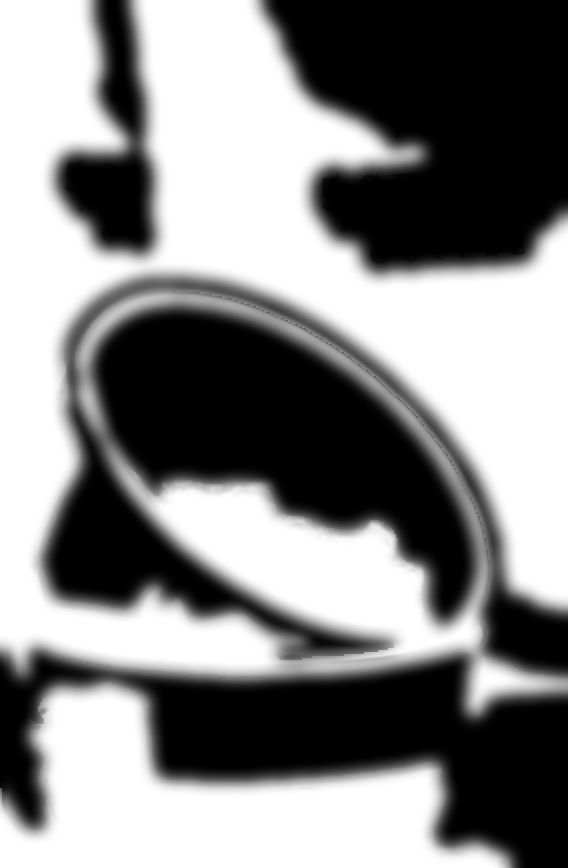}}
	\subfigure[{DPRL Map}]                    {\includegraphics[width=0.16\linewidth]{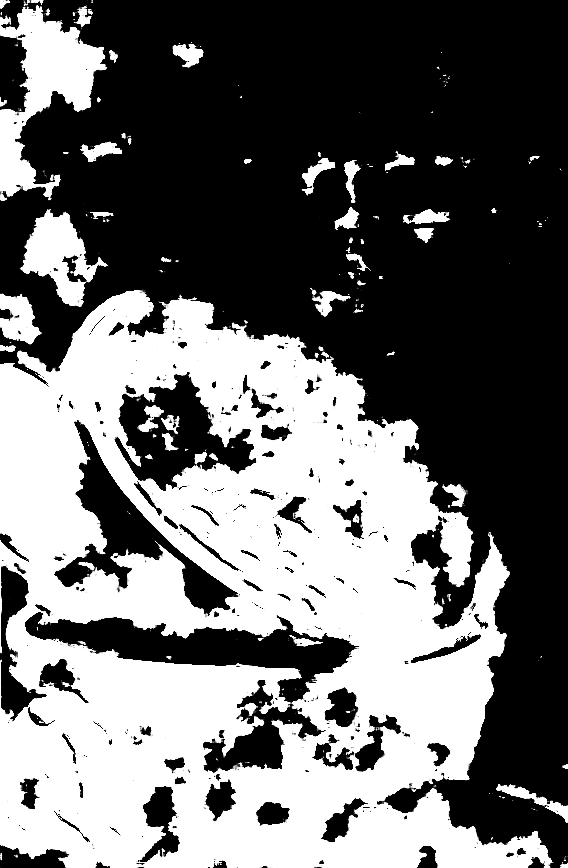}}
	\subfigure[{MMFNet Map}]                    {\includegraphics[width=0.16\linewidth]{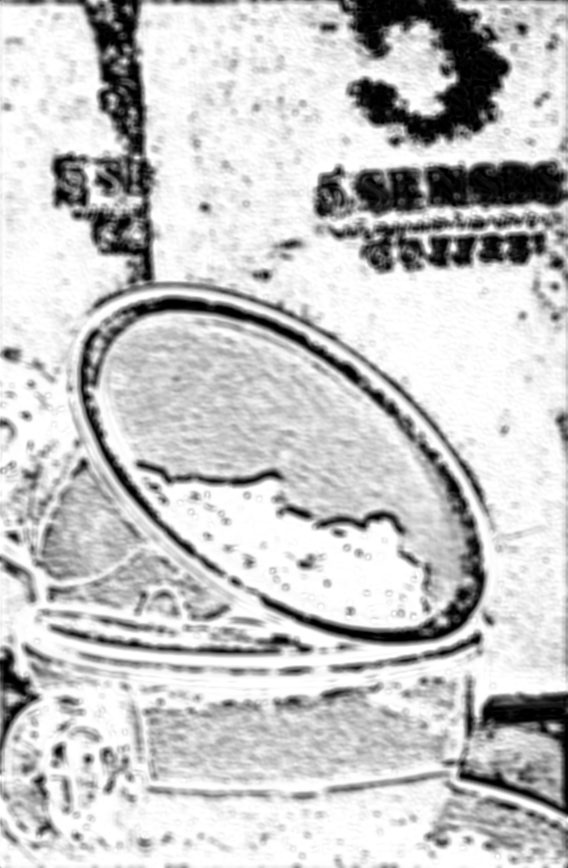}}
	\subfigure[{MS-Lap Map}]   {\includegraphics[width=0.16\linewidth]{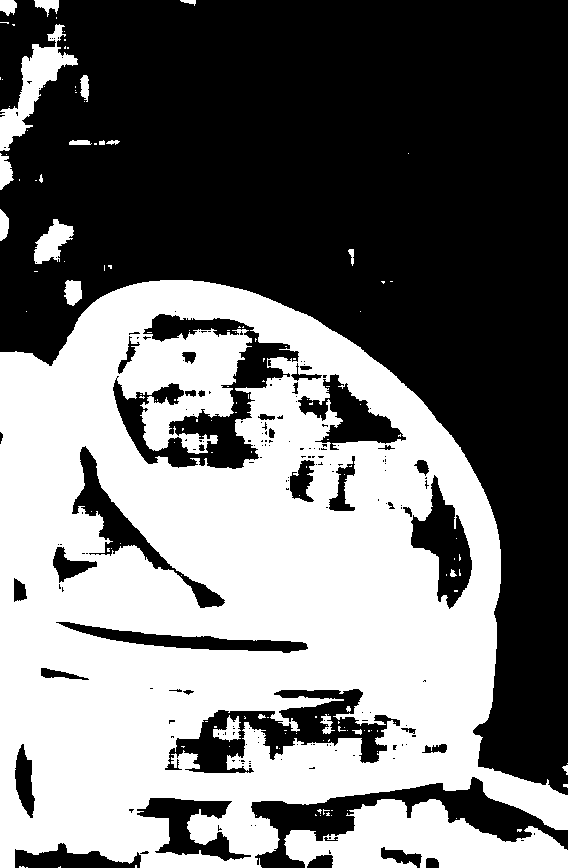}}
	\subfigure[{MS-ResNet Map}]         {\includegraphics[width=0.16\linewidth]{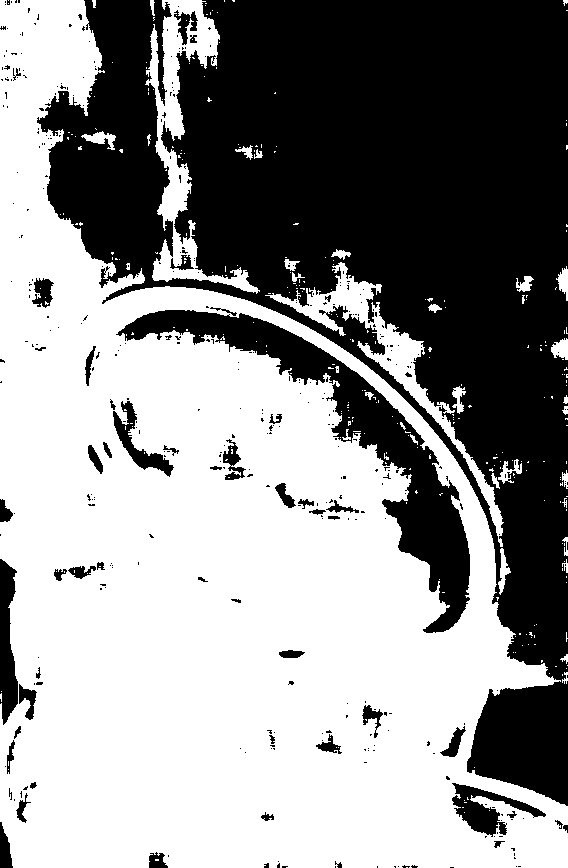}}

	\caption{(a-f): Fusion images. SSIM values are shown in the parentheses. (g-l): Detected focus maps. The reference and source images are displayed in Fig. \ref{fig:coffee_ori}. The manually annotated focus map is shown in Fig. \ref{fig:map-analysis}(b).} 
	\label{fig:coffee}
\end{figure*}

\section{Experiments}\label{sec:3}
We conduct extensive experiments on the real-world dataset to study behaviors and properties of the MFF-SSIM based fusion strategy. In what follows, our methods are abbreviated as MS-Lap and MS-ResNet. 

\subsection{Datasets and Metrics}
Our experiments' goal is to verify whether our framework does a better job than other methods if images suffer from severe defocus spread effects. In the following experiments, we employ the MFFW dataset \cite{MFFW} to evaluate the algorithms' perfromance. This dataset is presented recently and it contains 13 pairs of real-world multi-focus images collected on the Internet. This dataset provides annotated focus maps for each pair. In addition, to facilitate assessment, this dataset also releases the manually fused images. The scenes in MFFW are far more complicated and there is a significant defocus spread effect. It is a challenge to obtain satisfactory fusion images on this dataset. 

Normalized mutual information (NMI), Xydeas's metric \cite{Xydeas_metric} and Chen-Blum's metric \cite{Qcb} are employed as the no-reference based metrics. Since MFFW \cite{MFFW} provides the manually fused images, we use three reference image quality metrics to assess the algorithm performance, that is, peak signal-to-noise ratio (PSNR), SSIM \cite{SSIM} and feature similarity index (FSIM) \cite{FSIM}. Furthermore, except for these metrics we also report the mean opinion score (MOS). In detail, 10 volunteers were invited to evaluate the quality of fused images. All volunteers had no bias about this task. Their opinion score ranged from 1 to 10, and larger values corresponded to better images.

\subsection{Comparison with SOTA methods}\label{sec:3.2}
Our technique is compared with eight SOTA methods, including boundary founding (BF) \cite{BF}, BRW \cite{BRW}, CBF \cite{CBF}, GFF \cite{GFF}, CNN \cite{Liu_CNN}, DPRL \cite{DRPL}, ECNN \cite{ECNN} and MMFNet \cite{MMFNET}. The metrics are reported in Table \ref{tab:exp1}. The best and the second best values are highlighted by bold face and underline, respectively. It is shown that MS-ResNet achieves the highest PSNR, SSIM and FSIM values, and the second highest MOS value. MS-Lap has the best MOS value, and the second best PSNR, FSIM and Chen-Blum values. The no-reference based metrics show that our methods are comparable with popular counterparts as well. In the contrast, MMFNet almost performs worst, although the experiment in reference \cite{MMFNET} has proofed that MFF-Net outperforms others if the image suffers from mild defocus spread effects.

Besides the quantitative comparison, representative fusion images are visualized to further exhibit the effectiveness of MFF-SSIM based methods. The No. 6 image pair is displayed in Fig. \ref{fig:coffee_ori}. Two cups and two coffee bags are in near and distant focuses, respectively. Owing to the defocus spread effect, the characters on bags and the edges of two cups expand in source 1 and source 2 images, respectively. 
The fusion images and detected focus maps are displayed in Fig. \ref{fig:coffee}. It is shown that DPRL and MMFNet break down because they fragment the scene into many pieces and result in the obvious artifacts and ghosts. It indicates that their performances highly depend on the detector's accuracy. The fusion images generated by BF and CNN look more pleasant. However, for CNN, only the coffee beans and bags are clear, and most of the regions in two cups are still blurred; for BF, on account of defocus spread effects there are conspicuous haloes around the cups, so it cannot match up our expectations either. Obviously, MS-Lap and MS-ResNet generate the most satisfactory images, because all objects are clear and without artifacts or ghosts. At the same time, we can see that the focus maps detected by DPRL, MS-Lap and MS-ResNet are not accuracy enough, while DPRL fails in this case. The fact demonstrates that our proposed MFF-SSIM framework contributes more rather than the map detectors (i.e., the Laplacian energy and the ResNet).

According to the quantitative comparison and visual inspection, the conclusion can be drawn that MFF-SSIM based methods outperform others when the images suffer from defocus spread effects.
\begin{figure}[t]
	\centering
	\subfigure[Metric curves (versus corruption probability $p$)]  {\includegraphics[width=1\linewidth]{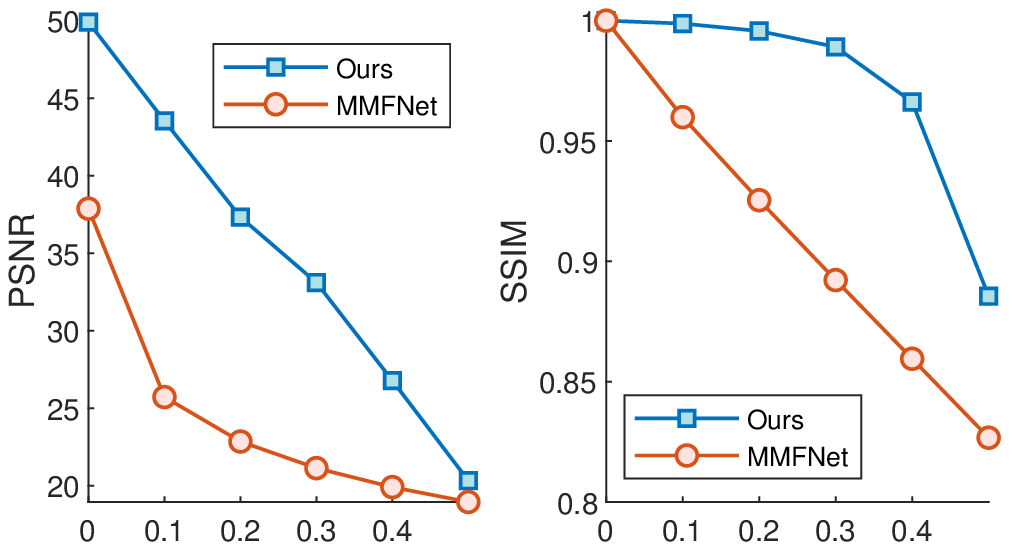}}
	\subfigure[{ Annotated Map}]  {\includegraphics[width=0.32\linewidth]{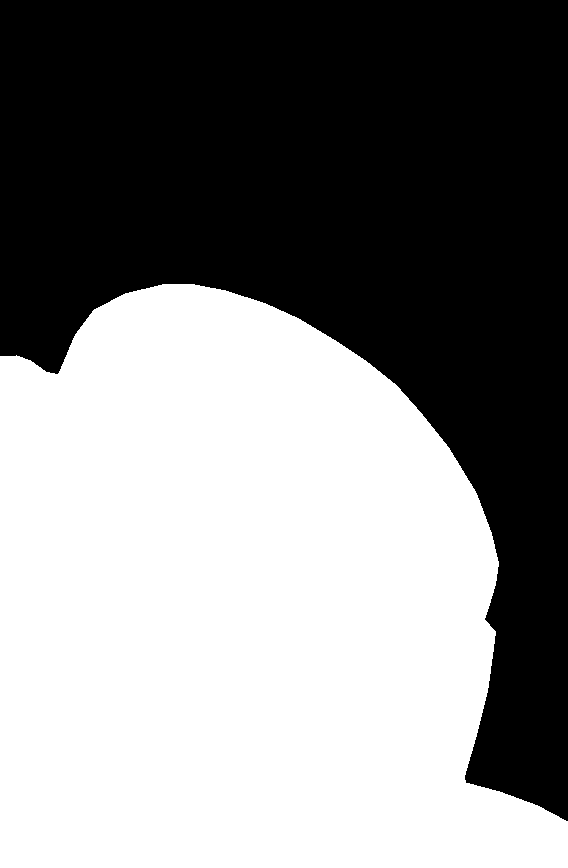}}
	\subfigure[{ Corrupted Map}]  {\includegraphics[width=0.32\linewidth]{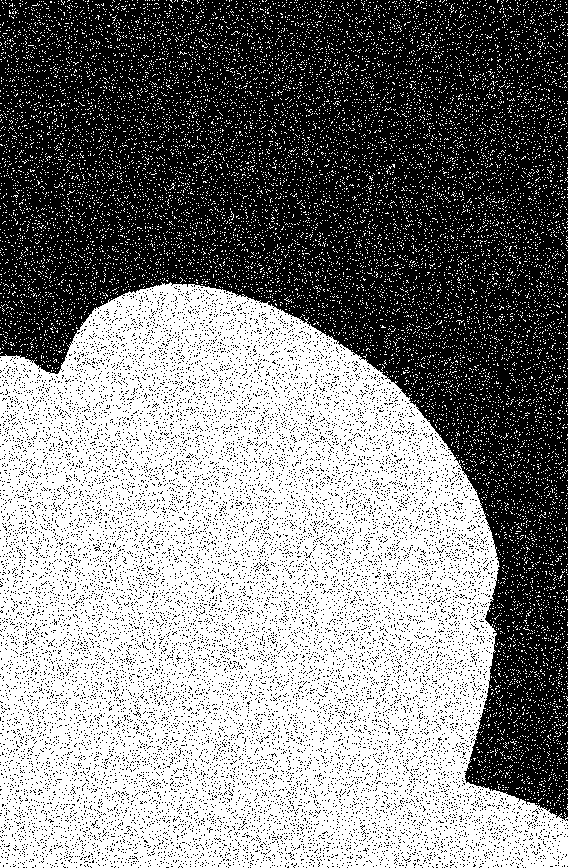}}
	\subfigure[{ Ours}]  {\includegraphics[width=0.32\linewidth]{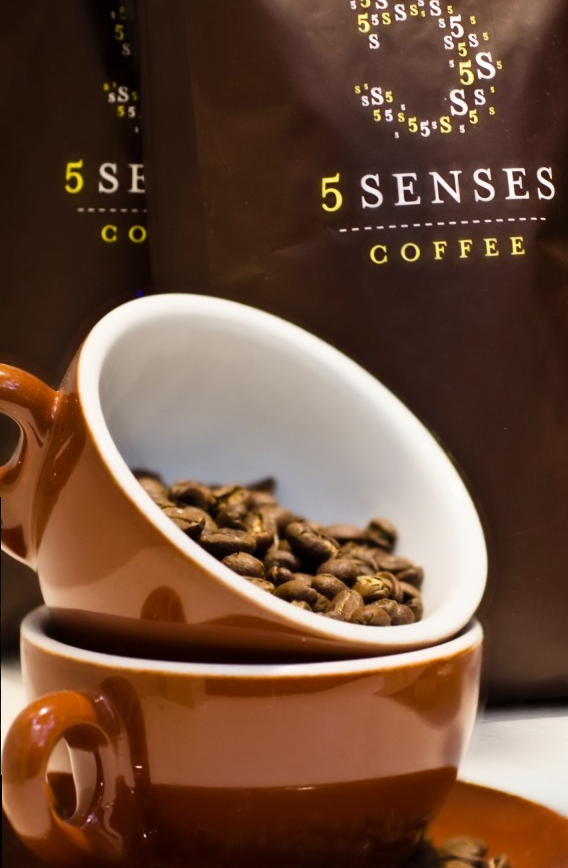}}
	\subfigure[{\scriptsize  Annotated Matte Map}]  {\includegraphics[width=0.32\linewidth]{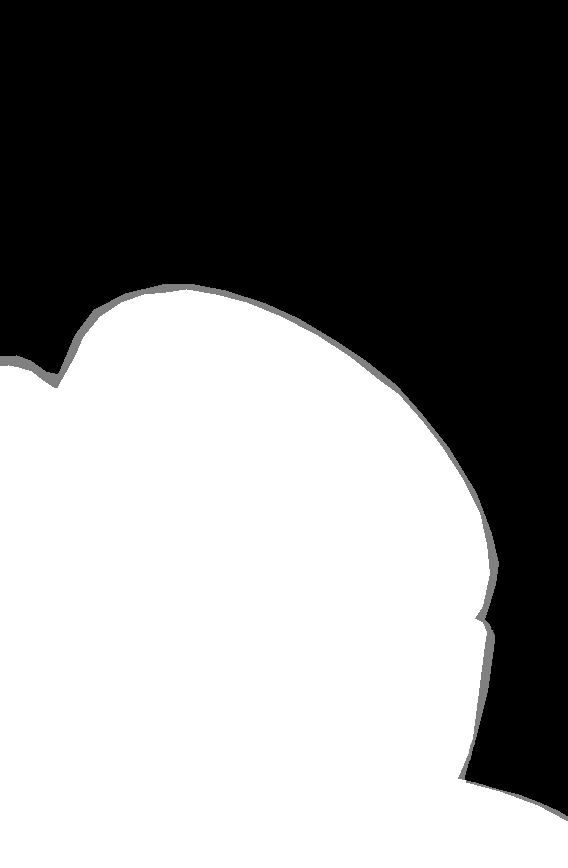}}
	\subfigure[{ Corrupted Matte Map}]  {\includegraphics[width=0.32\linewidth]{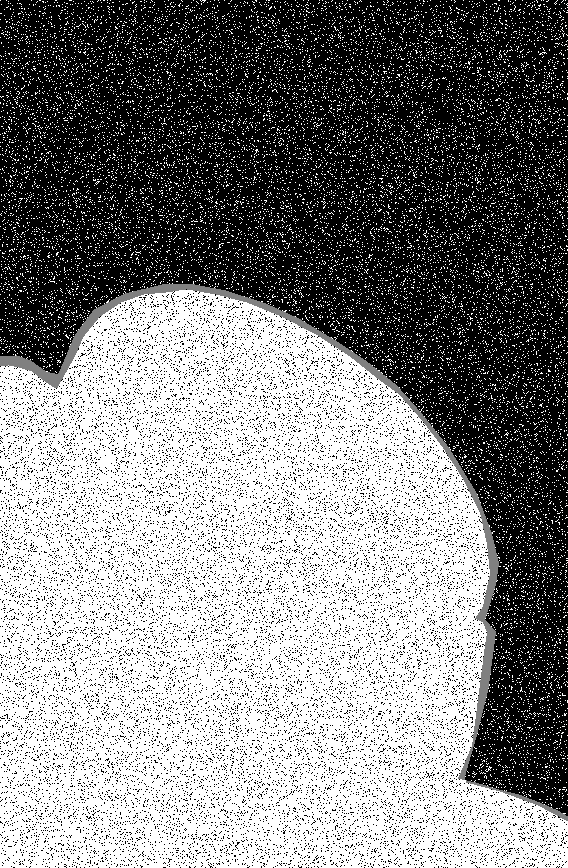}}
	\subfigure[{ MMFNet}]  {\includegraphics[width=0.32\linewidth]{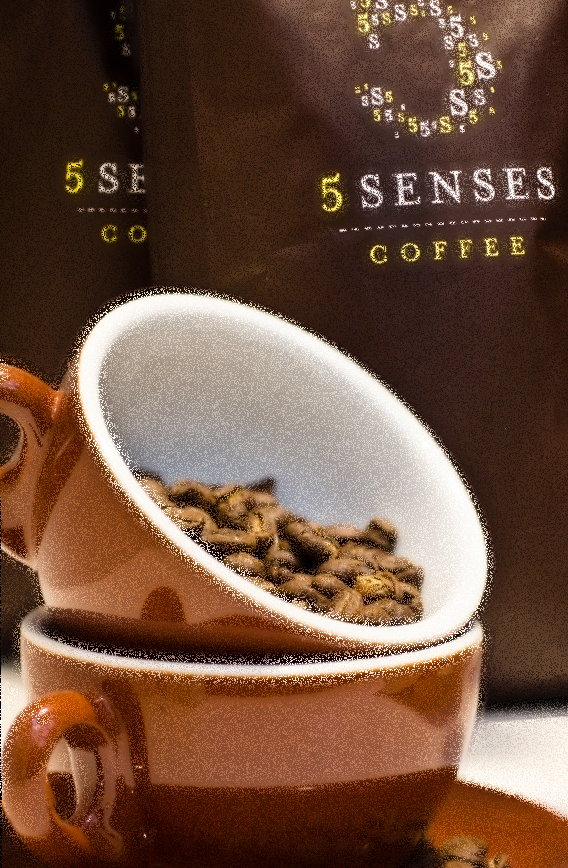}}
	\caption{Analysis on the focus map. (a) The metrics with different corruption probabilities. (b-d) The results of our method when $p=0.3$. (e-g) The results of MMF-Net when $p=0.3$. }
	\label{fig:map-analysis}
\end{figure}

\subsection{Robustness experiments}
Here, more experiments are conducted to analysis our proposed method.	
Firstly, we manually annotate the focus map, which is deemed to be accurate. Then, each pixel in this map is corrupted with a probability $p$. Note that corruption strategy is that the focused (defocused) pixel is changed to defocused (focused) one. Obivously the map is more inaccurate with greater $p$. At last, the corrupted map is taken as the input of MMF-SSIM fusion strategy. Similar steps are carried out for MMFNet. Our goals in this experiment are two-fold. The one is to observe how MFF-SSIM performs as the detected focus map getting inaccurate. The other one is to investigate whether MMF-SSIM fusion framework is more effective than the boundary post-processor in MMFNet.

This experiment is conducted 100 times, and the average PSNR/SSIM curves are displayed in Fig. \ref{fig:map-analysis}, where corruption probability $p$ ranges from 0 to 0.5 with a step 0.1. There is no doubt that the curves have downward tendencies as $p$ going greater, but the PSNR and SSIM values of MFF-SSIM model are always higher than those of MMFNet. In addition, we learn that when $p$ increases from 0 to 0.1, the PSNR value of MFF-SSIM model decreases from 49.91dB to 43.53dB by 12.78\%, while that of MMFNet dramatically decreases from 37.87dB to 25.72dB by 32.08\%. The fusion images with $p=0.3$ are also visualized in Fig. \ref{fig:map-analysis}. Our fusion image can match up our expectations. Nonetheless, the artifacts can be easily found in the image fused by MMFNet. Based on the above analysis, we can draw the conclusion that MFF-SSIM model is more robust than MMFNet.

\subsection{Ablation experiments}\label{sec:3.4}
In this subsection, a series of ablation experiments are conducted, that is, changing some hyper-parameters of the model and seeing how it affects performance. 
\subsubsection{Network depth and width}
\begin{figure}
	\centering
	\includegraphics[width=1\linewidth]{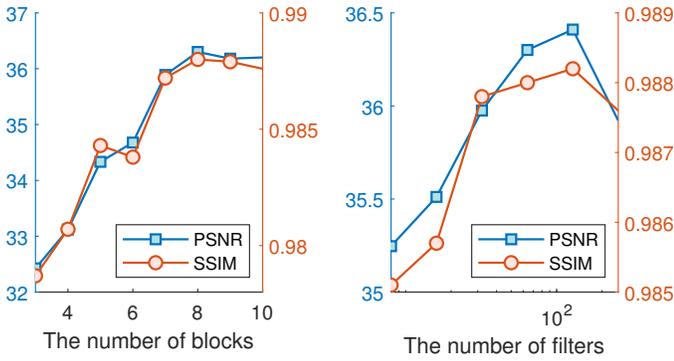}
	\caption{Effects of the network depth (left) and width (right).}
	\label{fig:ablationblocks}
\end{figure}

As shown in Fig. \ref{fig:structure2}, MS-ResNet employs 2 convolutional units and 8 residual blocks, and there are 128 filters except for the last convolutional unit. Here we analyze effects of network depth (that is, the number of residual blocks) and width (that is, the number of filters). The top panel of Fig. \ref{fig:ablationblocks} displays the PSNR and SSIM curves on the MFFW dataset with the number of residual blocks increasing from 3 to 10. It is shown that both PSNR and SSIM go larger and then tend to be flat with blocks' number growing. Eight blocks correspond to the best results. The bottom panel of Fig. \ref{fig:ablationblocks} shows the PSNR and SSIM curves with different numbers of filters, including 8, 16, 32, 64, 128 and 256. A similar conclusion can be drawn that both PSNR and SSIM go greater with filters' number growing. Nonetheless, it is found that larger blocks' or filters' number does not necessarily improve the performance of MS-ResNet. The reason may be that the MS-ResNet suffers from the overfitting problem when the network depth or width is too large. 

\subsubsection{Window size}
\begin{figure}[h]
	\centering
	\subfigure[{\scriptsize Source 1}]  {\includegraphics[width=0.30\linewidth]{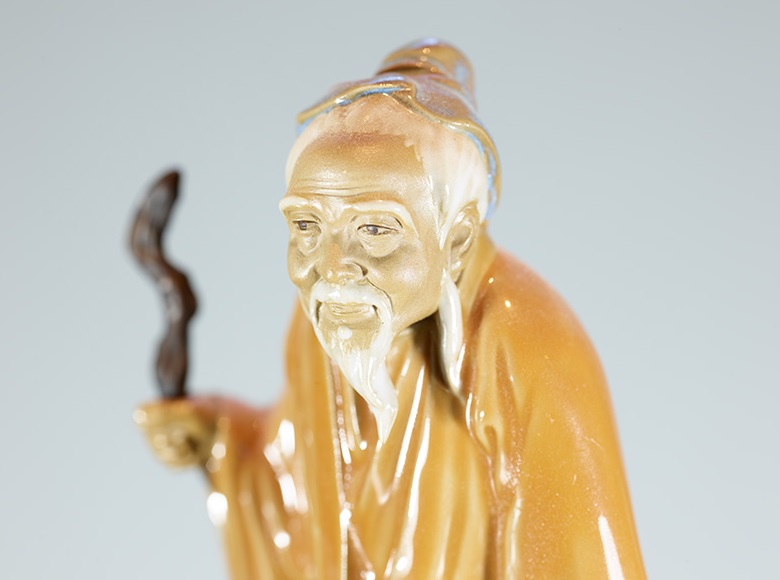}}
	\subfigure[{\scriptsize Source 2}]  {\includegraphics[width=0.30\linewidth]{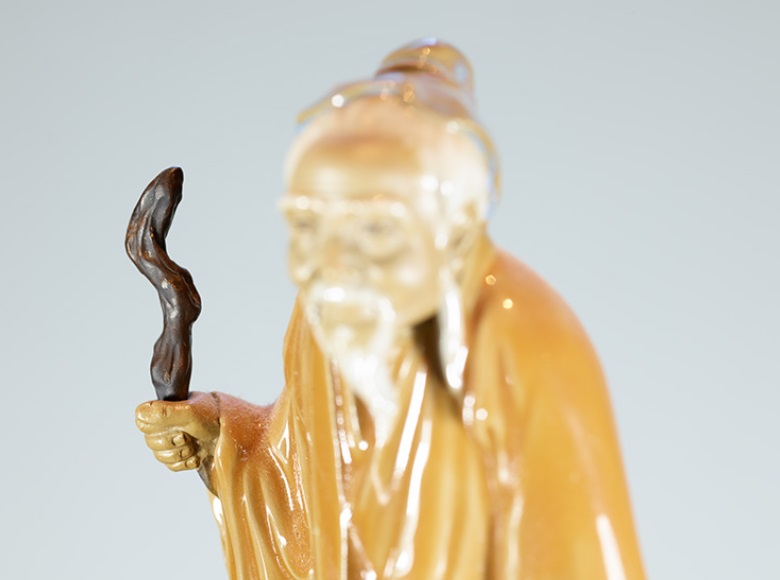}}
	\subfigure[{\scriptsize Reference}] {\includegraphics[width=0.30\linewidth]{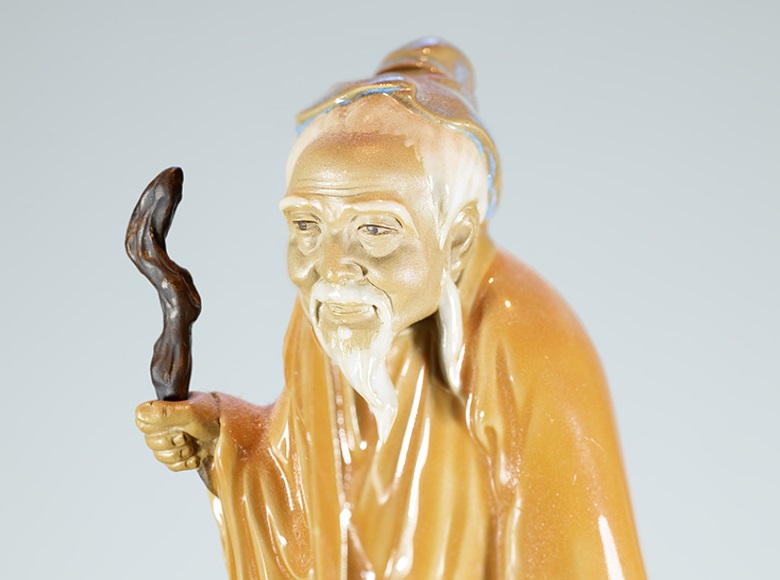}}
	\subfigure[{\scriptsize $\alpha =1.5\times10^{-5}$}] {\includegraphics[width=0.30\linewidth]{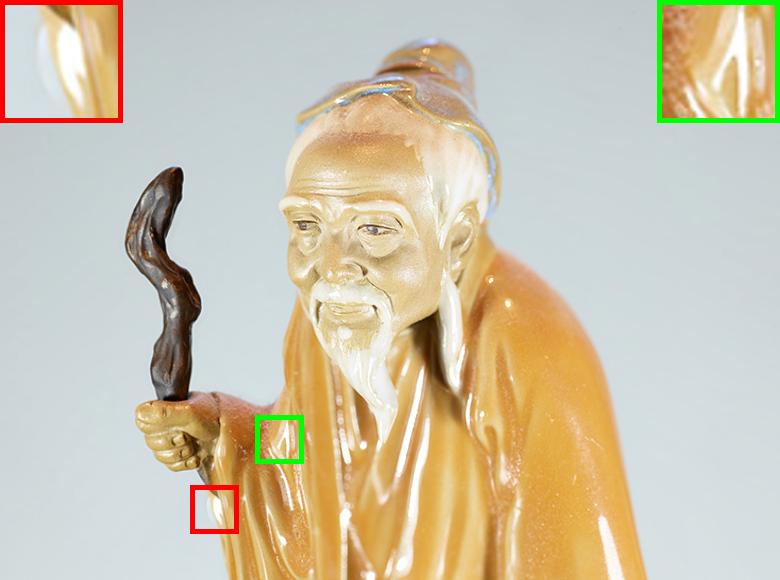}}
	\subfigure[{\scriptsize $\alpha =3.5\times10^{-5}$}] {\includegraphics[width=0.30\linewidth]{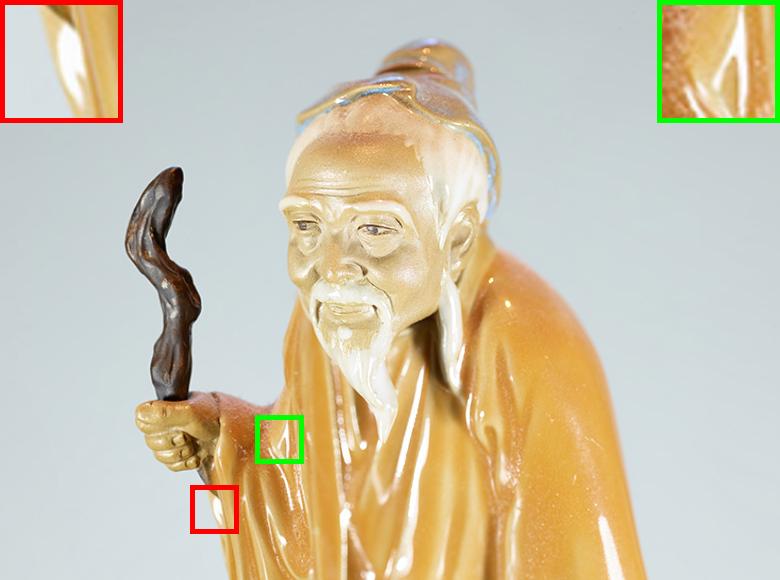}}
	\subfigure[{\scriptsize $\alpha =9.5\times10^{-5}$}] {\includegraphics[width=0.30\linewidth]{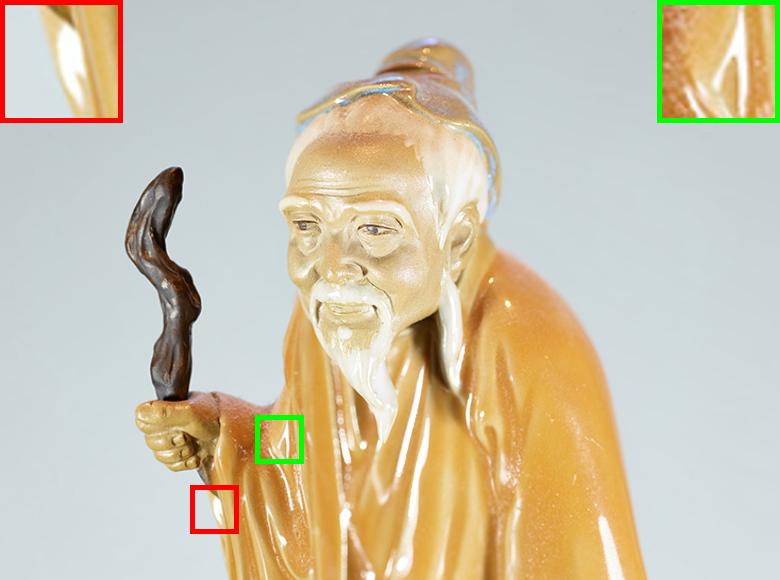}}
	\subfigure[{\scriptsize PSNR and SSIM curves}] {\includegraphics[width=\linewidth]{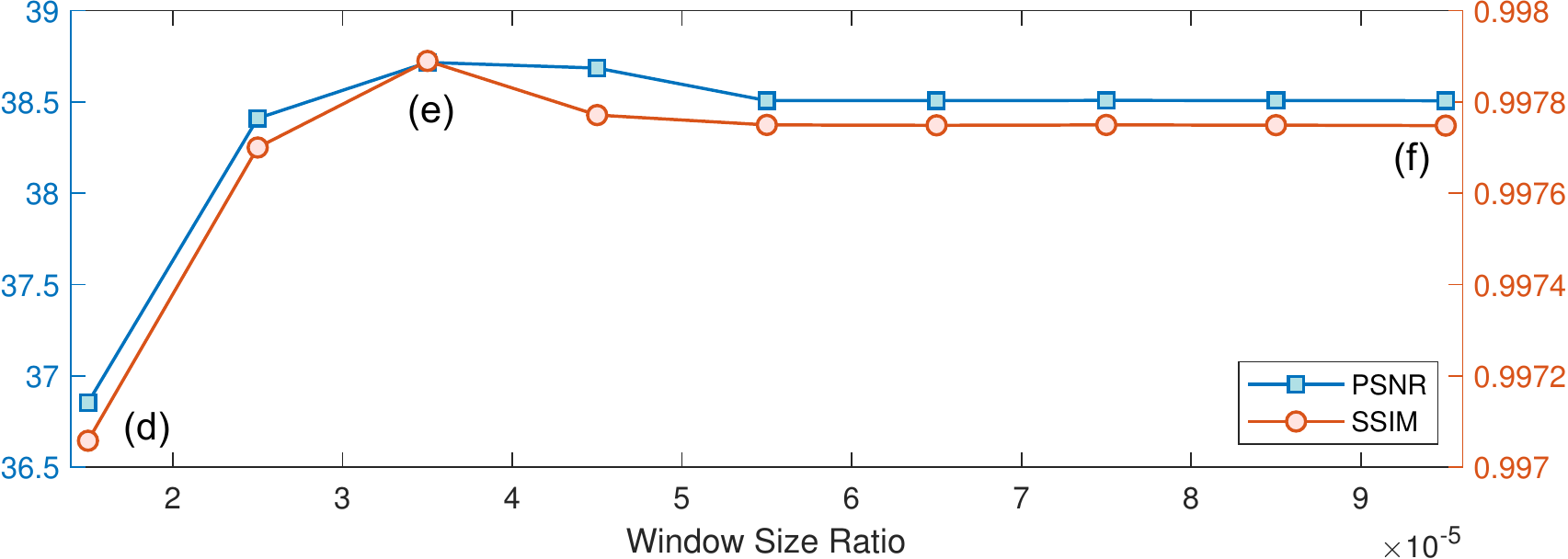}}
	\caption{The results with different window size ratios of No. 4 image set from MFFW.}
	\label{fig:pottery}
\end{figure}

\begin{figure}[t]
	\centering
	\subfigure[{\scriptsize Source 1}]  {\includegraphics[width=0.19\linewidth]{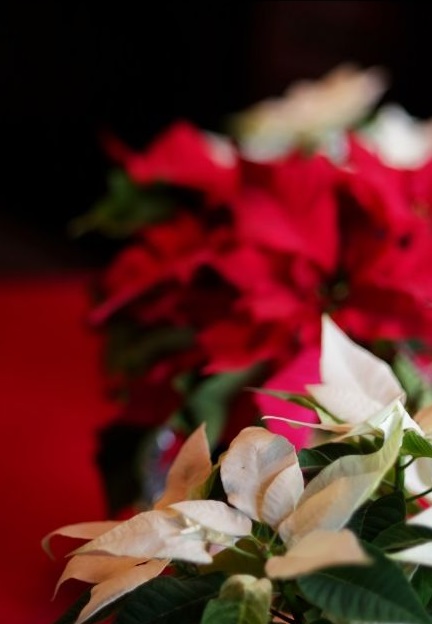}}
	\subfigure[{\scriptsize Source 2}]  {\includegraphics[width=0.19\linewidth]{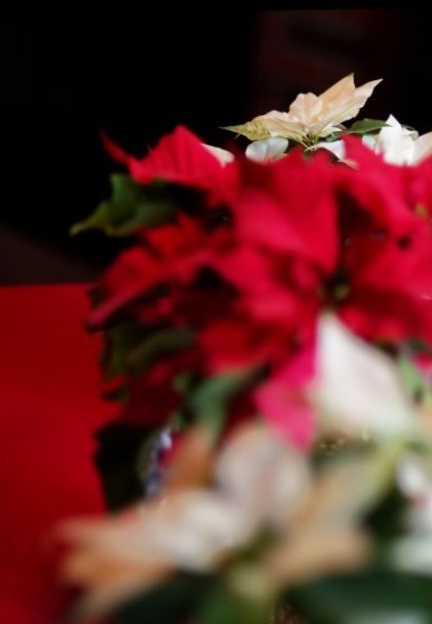}}
	\subfigure[{\scriptsize Reference}] {\includegraphics[width=0.19\linewidth]{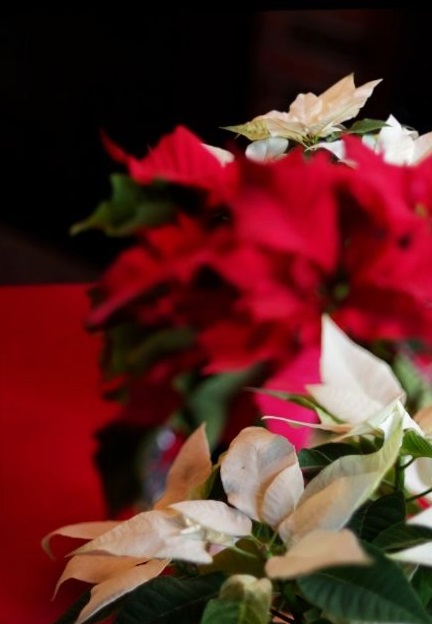}}
	\subfigure[{\scriptsize $\alpha =1.5\times10^{-5}$}] {\includegraphics[width=0.19\linewidth]{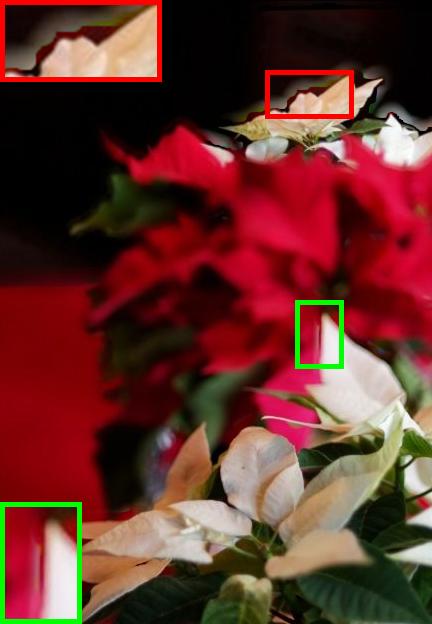}}
	\subfigure[{\scriptsize $\alpha =9.5\times10^{-5}$}] {\includegraphics[width=0.19\linewidth]{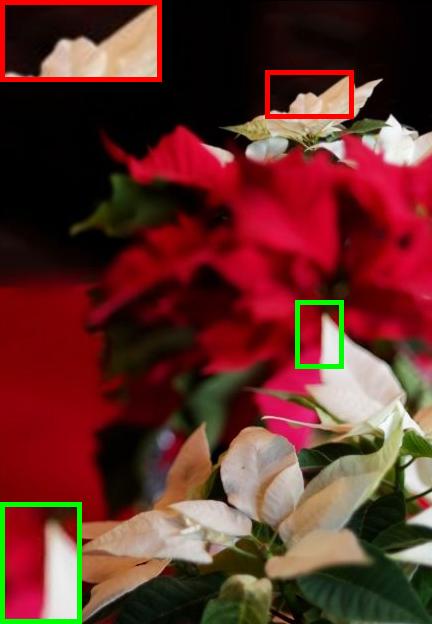}}
	\subfigure[{\scriptsize PSNR and SSIM curves}] {\includegraphics[width=\linewidth]{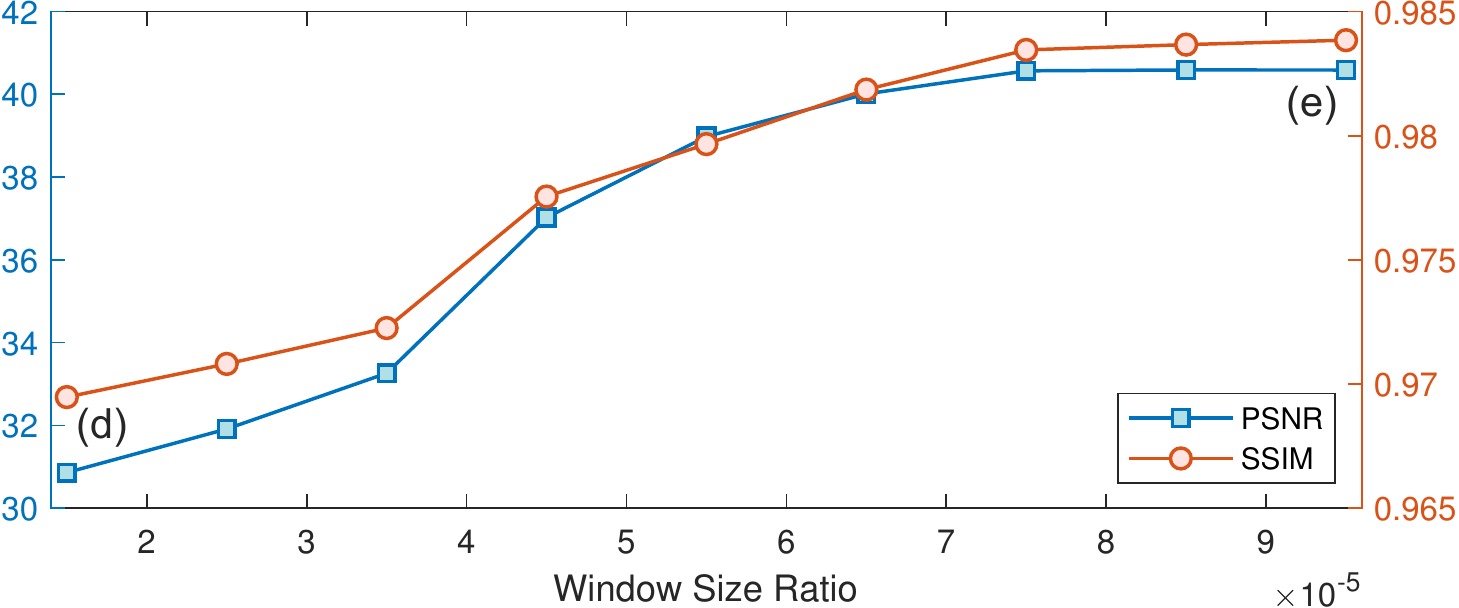}}
	\caption{The results with different window size ratios of No. 11 image set from MFFW.}
	\label{fig:rose}
\end{figure}

In above experiments, the window size is empirically set as $W=\alpha MN$, where the window size ratio is set as $\alpha =5\times10^{-5}$, and $M$ and $N$ denote the height and width of the image, respectively. Generally speaking, the configuration of window size is important to patch-wise methods. In this experiment, it aims at investigating the performance of MFF-SSIM-Strategy with different window size ratios. For simplicity, the LAP is employed as the focus map detector, and two pairs of images from MFFW are taken as representative examples. 

The window size ratio is sampled from $1.5\times10^{-5}$ to $9.5\times10^{-5}$ with step $1\times10^{-5}$. Figs. \ref{fig:pottery} and \ref{fig:rose} display the PSNR and SSIM curves, and the fusion images with the smallest, largest and optimal window sizes. In the first example, the defocus spread effect is relatively mild. It is found that as $\alpha $ increasing both PSNR and SSIM get larger and then decrease. The best result corresponds to $\alpha =3.5\times10^{-5}$. As for visual inspection, Fig. \ref{fig:pottery} (d) reveals that small window size suffers from artifacts. In addition, it is observed that the fusion image with large window size (see Fig. \ref{fig:pottery} (f)) is visually similar to optimal fusion image (see Fig. \ref{fig:pottery} (e)). In the second example, the defocus spread effect is relatively severe. From Fig. \ref{fig:rose}, it is learned that both PSNR and SSIM increase as window size increasing, and the best result is reached at $\alpha =9.5\times10^{-5}$. And the fusion image with larger window size is visually better than that with smaller window size. In summary, larger the window size is, better our method is. This conclusion matches up our anticipation, to some degree, because larger window size indicates that there are more neighbors help the center point to determine its pixel value. 

\begin{figure}
	\centering
	\subfigure[{\scriptsize Source 1}]  {\includegraphics[width=0.48\linewidth]{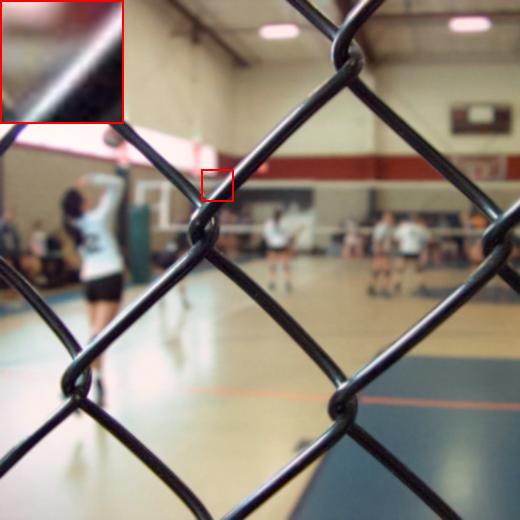}}
	\subfigure[{\scriptsize Source 2}]  {\includegraphics[width=0.48\linewidth]{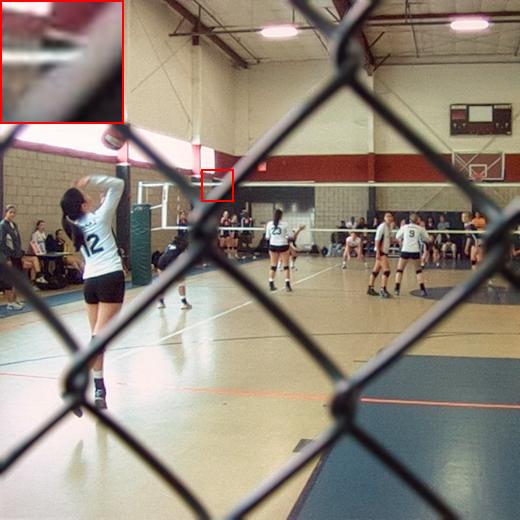}}
	\subfigure[{\scriptsize Local patches}] {\includegraphics[width=1\linewidth]{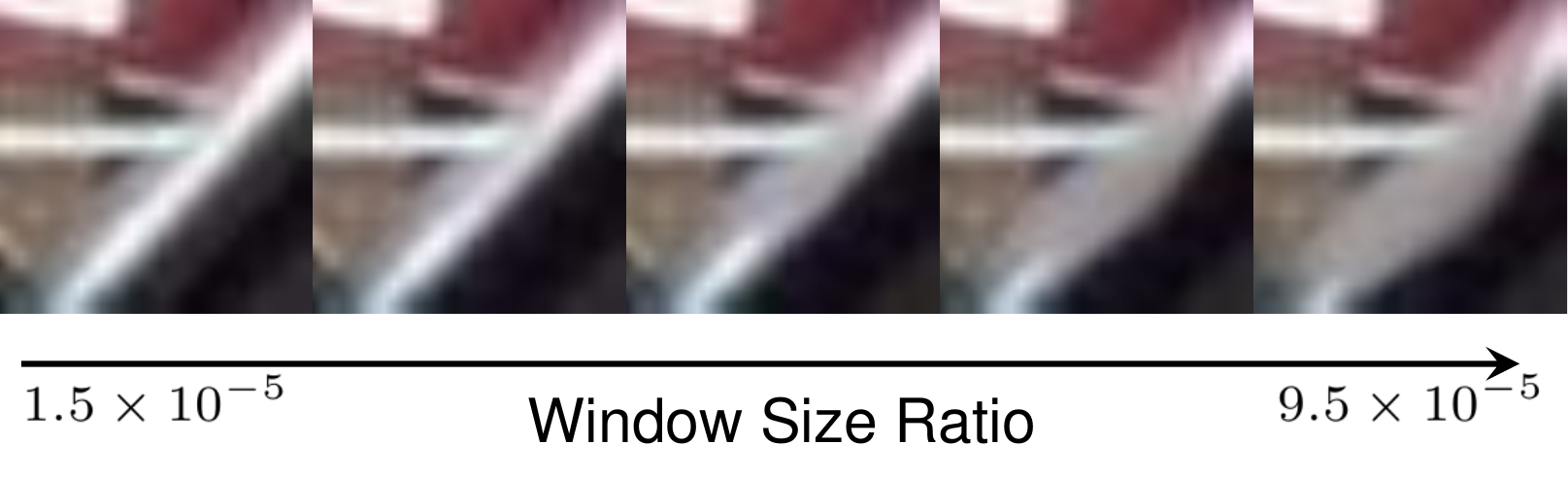}}
	\caption{The results with different window size ratios of No. 5 image set from Lytro. Since there is no ground truth, we only display the concerned local patches instead of the PSNR and SSIM curves.}
	\label{fig:fence}
\end{figure}

However, this conclusion will not stay true if the foreground or background is disconnected. For example, as shown in Fig. \ref{fig:fence} larger window size would do harm to the fusion image, when there is a crossed fence. In addition, it is worthy pointing out that larger window size significantly raises execution time. It still remains a problem that how to automatically pick optimal window size so as to strike the balance between performance and running speed.  

\subsection{Multiple source images}
\begin{figure}
	\centering
	\centering
	\subfigure[{Source 1}]          {\includegraphics[width=0.32\linewidth]{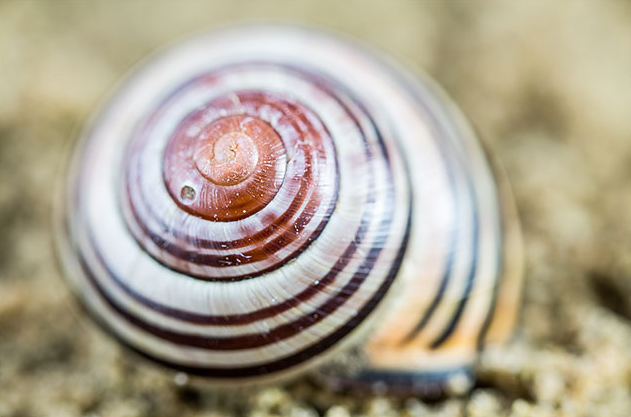}}
	\subfigure[{Source 2}]          {\includegraphics[width=0.32\linewidth]{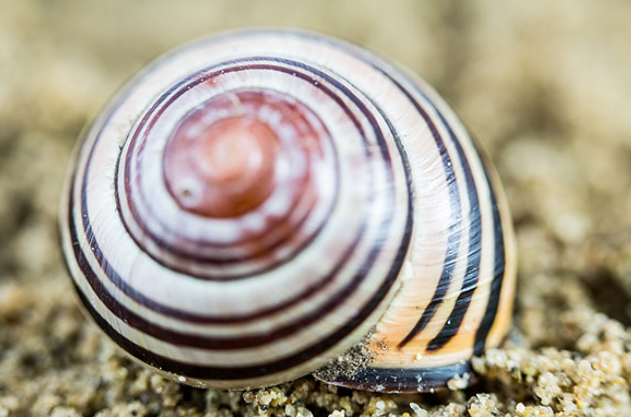}}
	\subfigure[{Source 3}]          {\includegraphics[width=0.32\linewidth]{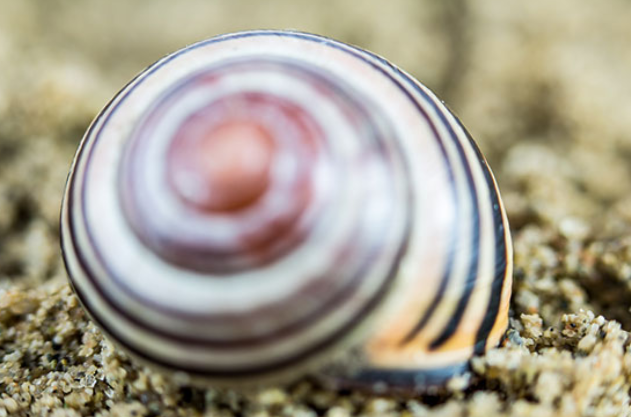}}
	\subfigure[{BF}]                {\includegraphics[width=0.32\linewidth]{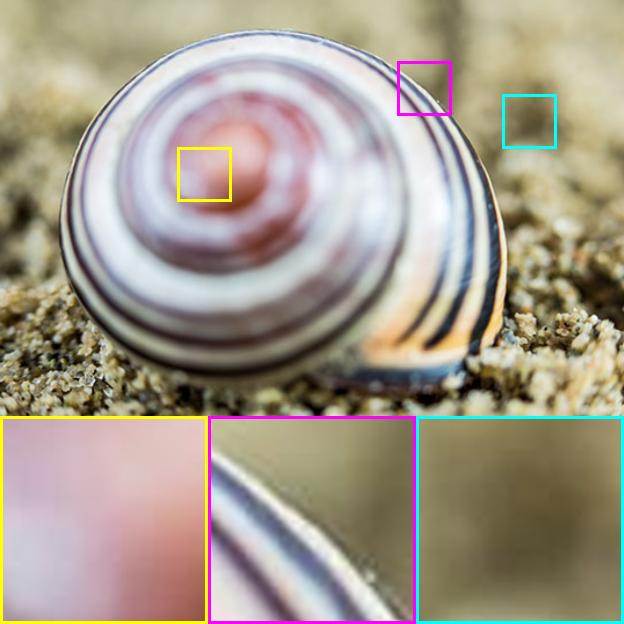}}
	\subfigure[{CNN}]               {\includegraphics[width=0.32\linewidth]{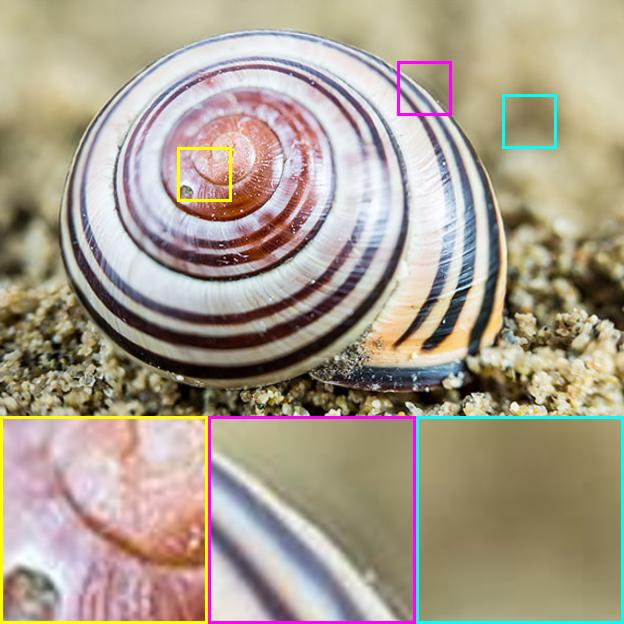}}
	\subfigure[{DPRL}]              {\includegraphics[width=0.32\linewidth]{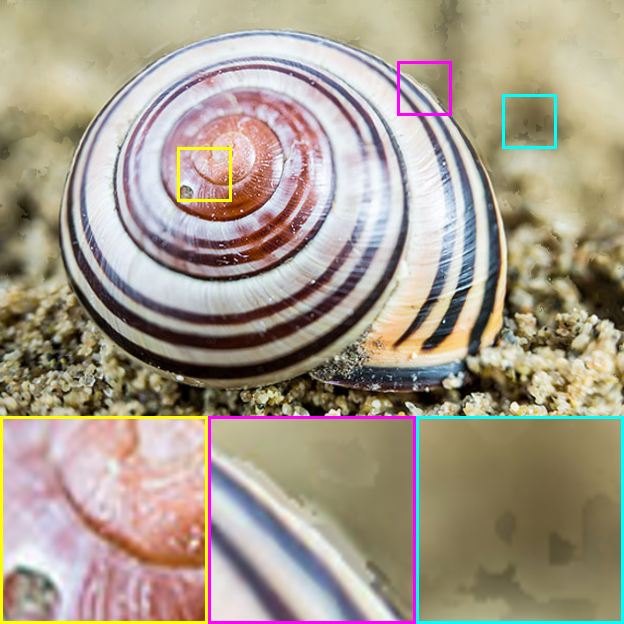}}
	\subfigure[{MMFNet}]            {\includegraphics[width=0.32\linewidth]{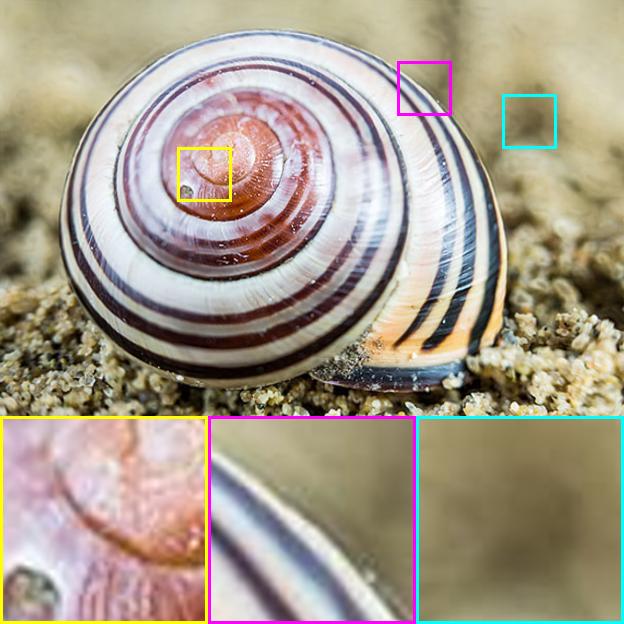}}
	\subfigure[{MS-Lap }]           {\includegraphics[width=0.32\linewidth]{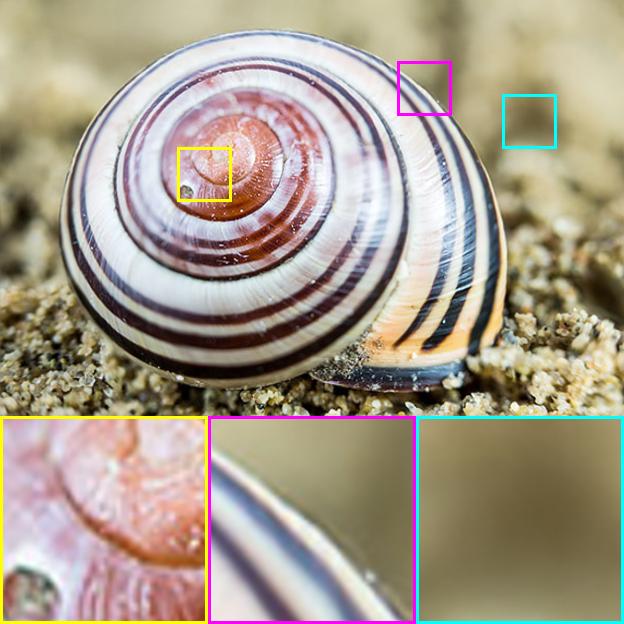}}
	\subfigure[{MS-ResNet}]         {\includegraphics[width=0.32\linewidth]{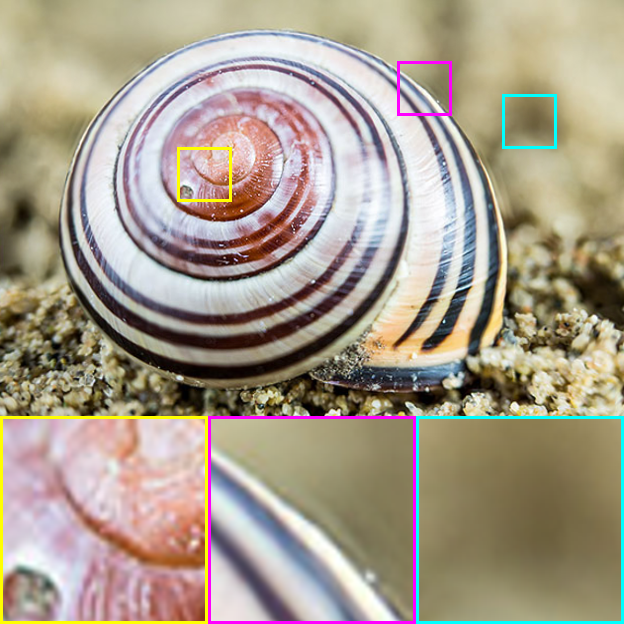}}
	\caption{Fusion results of a set of multiple source images.}
	\label{fig:shell}
\end{figure}

The above experiments mainly focus on the case of two source images. It is interesting to study the performance of our method if there are multiple images (i.e., $K>2$). Recall the updating rule, viz. Eq. (\ref{eq:grdt}). We know that if the focus map detector can generate the maps for $K$ source images simultaneously (e.g. the Laplace energy method), MFF-SSIM model will be directly applied in this case. However, if the focus map detector only deals with two source images (e.g. ResNet), we have to fuse them one by one. A representative image set is displayed in Fig. \ref{fig:shell}. There is a shell, whose different parts are in the near, middle and distant focuses. It is shown that our methods still provide satisfactory images.

\subsection{Mild defocus spread effect experiments}
Though the mild defocus spread effect is out of our scope, it is also interesting to investigate the behavior of our methods in this case. To this end, the algorithms are applied to the Lytro and Grayscale datasets who suffer from mild defocus spread effects. With the limitation of paper length, the results are displayed in supplementary materials. Although our methods are not the best performer on Lytro and Grayscale datasets, our fusion images are artifact-free and are visually similar to the best one (i.e., MMFNet).


\section{Conclusions}\label{sec:4}
This paper presents an SSIM-based multi-focus image fusion framework, the first attempts to deal with severe defocus spread effects. The experimental results show that our framework outperforms the SOTA methods. Our fusion images are artifact-free, while others contains obvious artifacts. However, our method is time-consuming. It is interesting to investigate how to design a real-time algorithm to deal with defocus spread effects in the future. And, light filed imaging provides the richer visual information than the classic photography and it has been applied to depth estimation and super-resolution \cite{DBLP:journals/tip/ShengZCFX17,DBLP:journals/tip/WuLDC19}. Therefore, another interesting future work is how to effectively combine multi-focus images with the light field data to overcome the defocus spread effects.


%

\ifCLASSOPTIONcaptionsoff
  \newpage
\fi



%

\bibliographystyle{IEEEtran}
\bibliography{egbib}
%




\end{document}